\documentclass{article}

\usepackage{microtype}
\usepackage{graphicx}
\usepackage{subfigure}
\usepackage{booktabs}
\usepackage{tikz}
\usepackage{standalone}
\usetikzlibrary{arrows.meta,positioning,calc,fit,shapes.geometric}
\usepackage{hyperref}

\newif\ificmlpreprint
\icmlpreprinttrue
\ificmlpreprint
    \usepackage[accepted]{icml2025}
    \makeatletter
    \renewcommand{\ICML@appearing}{Preprint.}
    \makeatother
    \hypersetup{pdfsubject={Preprint},hypertexnames=false}
\else
    \usepackage{icml2025}
\fi

\usepackage{amsmath}
\usepackage{amssymb}
\usepackage{mathtools}
\usepackage{amsthm}
\providecommand{\vct}[1]{\boldsymbol{#1}}
\providecommand{\mat}[1]{\boldsymbol{#1}}
\usepackage[capitalize,noabbrev]{cleveref}

\renewcommand{\algorithmicrequire}{\textbf{Input:}}
\renewcommand{\algorithmicensure}{\textbf{Output:}}

\theoremstyle{plain}
\newtheorem{theorem}{Theorem}[section]

\newtheorem{lemma}[theorem]{Lemma}

\theoremstyle{definition}

\theoremstyle{remark}

\icmltitlerunning{Dual DeltaNet}

\begin{document}

\twocolumn[
\icmltitle{Dual DeltaNet: Value-Side Memory Reorganization for Linear Attention}

\icmlsetsymbol{corresponding}{*}
\begin{icmlauthorlist}
\icmlauthor{Yixiao Qian}{zju}
\icmlauthor{Yaxin Zhang}{zjupi}
\icmlauthor{Song Chen}{nus,corresponding}
\icmlauthor{Jiaxu Liu}{huzhou}
\icmlauthor{Shengze Cai}{zju,corresponding}
\icmlauthor{Chao Xu}{zju,corresponding}
\end{icmlauthorlist}

\icmlaffiliation{zju}{College of Control Science and Engineering, Zhejiang University}
\icmlaffiliation{zjupi}{Polytechnic Institute, Zhejiang University}
\icmlaffiliation{nus}{Department of Mathematics, National University of Singapore}
\icmlaffiliation{huzhou}{School of Science, Huzhou Normal University}
\icmlcorrespondingauthor{Song Chen}{song.chen@nus.edu.sg}
\icmlcorrespondingauthor{Shengze Cai}{shengze\_cai@zju.edu.cn}
\icmlcorrespondingauthor{Chao Xu}{cxu@zju.edu.cn}
\icmlkeywords{Linear Attention, Recurrent Models, Gated DeltaNet, Long-Context Retrieval}

\vskip 0.3in
]

\printAffiliationsAndNotice{}

\begin{abstract}
Linear attention and state space models (SSMs) offer efficient alternatives 
to softmax attention by maintaining a fixed-size recurrent state,
but this design requires careful management 
to preserve useful historical associations and limit interference.
The delta rule helps reduce interference through correction of the value 
associated with each key, but it does not explicitly account for 
how related value features are distributed across the state.
We observe that a state $\boldsymbol{S}\in\mathbb{R}^{d_k\times d_v}$ 
can be viewed as a bank of $d_k$ addresses storing $d_v$-dimensional values. 
While a key-side read $\boldsymbol{S}^\top\boldsymbol{k}$ retrieves the value 
associated with $\boldsymbol{k}$, a value-side read $\boldsymbol{S}\boldsymbol{e}$ aggregates 
addresses whose stored value features match the selector $\boldsymbol{e}$.
Based on this observation, we propose Dual DeltaNet (DDN), which extends Kimi
Delta Attention (KDA) with a two-sided delta rule.
The key-side delta rule reads with the current key to correct its associated
value, while the complementary value-side delta rule reads with selected
features of the current value to reorganize the addresses it occupies.
We derive a chunkwise parallel training algorithm and a rank-2 recurrence 
that increases arithmetic intensity during inference. 
Compared with KDA, GDN-2, and Mamba-3, 
DDN achieves the strongest overall performance across 
language modeling, zero-shot downstream tasks, and retrieval, 
while incurring modest training and inference overhead relative to KDA.
\textbf{Code:} \url{https://github.com/ReichtumQian/DualDeltaNet}
\end{abstract}

\section{Introduction}

Transformers have become the dominant backbone of modern large language
models~\citep{vaswani2017attention}.
However, full attention incurs quadratic computation in sequence length
during training, while autoregressive inference maintains a token-level KV
cache that grows linearly with the context length.
Modern recurrent sequence models, including state space models (SSMs) and
linear attention, offer efficient alternatives by processing sequences through
recurrent state updates and compressing the history into a fixed-size
state~\citep{gu2023mamba,dao2024transformers,qian2026dssms,katharopoulos2020transformers}.
This recurrent formulation enables linear-time sequence processing and
fixed-size inference memory, and recent work has shown that such models can
achieve language-modeling performance competitive with
Transformers~\citep{yang2025gated,team2025kimi,lahoti2026mamba3}.
This compression, however, introduces a fundamental challenge:
the recurrent fixed-size state must continuously accommodate new information
while preserving useful historical associations and avoiding interference among them.
Prior studies on associative recall, copying, and long-context retrieval
suggest that recurrent models can be less effective at 
retrieving precise information than softmax attention in certain settings.
~\citep{arora2024zoology,jelassi2024repeat,waleffe2024empirical}.
Therefore, effectively utilizing the recurrent fixed-size state while maintaining reliable
associative memory remains a central challenge.

Prior work has sought to improve recurrent-state utilization along several
complementary directions. The Mamba family improves the underlying state-space
dynamics:
Mamba introduces input-dependent selective propagation and forgetting,
Mamba-2 develops state space duality and a more efficient structured
recurrence, and Mamba-3 combines a more expressive discretization with
complex-valued states and multi-input multi-output updates
\citep{gu2023mamba,dao2024transformers,lahoti2026mamba3}. In parallel, the
DeltaNet family improves the associative update itself. DeltaNet uses the
delta rule to correct the value associated with the current key
\citep{schlag2021linear}, and a subsequent parallel formulation enables
efficient training over sequence length~\citep{yang2024parallelizing}.
Gated DeltaNet (GDN) adds an adaptive forgetting gate,
Kimi Delta Attention (KDA) extends decay to individual key
channels, and Gated DeltaNet-2 (GDN-2) further decouples erasure from writing
\citep{yang2025gated,team2025kimi,hatamizadeh2026gated}.
Preconditioned GDN (PGDN) and Kalman Delta Networks (KDNs) further refine
residual writes using curvature and uncertainty information, respectively
\citep{tumma2026preconditioned,bui2026kalman}.
Another line organizes memory into explicit slots with separately maintained
key and value states. Gated Slot Attention (GSA) uses input-dependent routing and adaptive forgetting
to update these slots, Raven sparsifies this routing so that each token
modifies only a selected subset of slots, and GSA2 implements two-sided
correction through separate key--slot and slot--value recurrences, using
Gated Oja Rule-2 and Gated Delta Rule-2, respectively
\citep{zhang2024gatedslot,afzal2026raven,li2026gatedslot2}. These approaches
improve recurrent memory through richer dynamics, associative correction, or
explicit slot routing.

\begin{figure*}[t]
    \centering
    \resizebox{0.8\textwidth}{!}{%
        \begingroup
        \endlinechar=-1
        \definecolor{currentblue}{HTML}{5576C2}
\definecolor{selectorgreen}{HTML}{6AA84F}
\definecolor{targetorange}{HTML}{E67E22}
\definecolor{residualpurple}{HTML}{8E63B0}
\definecolor{stategray}{HTML}{B8B8B8}
\definecolor{panelgray}{HTML}{F7F7F8}

\pgfmathsetmacro{\respzero}{.10*0+.95*.80+.10*0+.90*.60+.10*0}
\pgfmathsetmacro{\respone}{.10*0+0*.80+.15*0+0*.60+.10*0}
\pgfmathsetmacro{\resptwo}{.10*0+.80*.80+.10*0+.75*.60+.10*0}
\pgfmathsetmacro{\respthree}{.10*0+0*.80+.10*0+0*.60+.10*0}
\pgfmathsetmacro{\respfour}{.10*0+0*.80+.10*0+0*.60+.10*0}
\pgfmathsetmacro{\respmax}{\respzero}

\pgfmathsetmacro{\curzero}{\respzero/\respmax}
\pgfmathsetmacro{\curone}{\respone/\respmax}
\pgfmathsetmacro{\curtwo}{\resptwo/\respmax}
\pgfmathsetmacro{\curthree}{\respthree/\respmax}
\pgfmathsetmacro{\curfour}{\respfour/\respmax}

\pgfmathsetmacro{\tarzero}{0}
\pgfmathsetmacro{\tarone}{0}
\pgfmathsetmacro{\tartwo}{0}
\pgfmathsetmacro{\tarthree}{0}
\pgfmathsetmacro{\tarfour}{1}
\pgfmathsetmacro{\czero}{\tarzero-\curzero}
\pgfmathsetmacro{\cone}{\tarone-\curone}
\pgfmathsetmacro{\ctwo}{\tartwo-\curtwo}
\pgfmathsetmacro{\cthree}{\tarthree-\curthree}
\pgfmathsetmacro{\cfour}{\tarfour-\curfour}

\pgfmathsetmacro{\newzero}{\curzero+.85*\czero}
\pgfmathsetmacro{\newone}{\curone+.85*\cone}
\pgfmathsetmacro{\newtwo}{\curtwo+.85*\ctwo}
\pgfmathsetmacro{\newthree}{\curthree+.85*\cthree}
\pgfmathsetmacro{\newfour}{\curfour+.85*\cfour}

\begin{tikzpicture}[
    font=\small,
    panel/.style={
        draw=black!45,
        rounded corners=3pt,
        line width=0.8pt,
        fill=panelgray,
        minimum width=7.80cm,
        minimum height=4.95cm
    },
    paneltitle/.style={font=\small\bfseries, text=black!88},
    label/.style={font=\scriptsize, text=black!72, align=center},
    tinylabel/.style={font=\tiny, text=black!65, align=center},
    toplabel/.style={
        font=\scriptsize,
        text=black!72,
        anchor=base,
        text height=1.6ex,
        text depth=0.3ex
    },
    mathlabel/.style={
        font=\scriptsize,
        anchor=base,
        text height=2.0ex,
        text depth=0.7ex
    },
    flow/.style={-{Stealth[length=4pt,width=5pt]}, line width=0.9pt},
    cell/.style={draw=white, line width=0.4pt}
]

\node[panel] (panelA) at (3.90,2.55) {};
\node[paneltitle, anchor=north] at ([yshift=-0.20cm]panelA.north)
    {(a) Value-side read};

\coordinate (SA) at (1.00,1.40);
\foreach \r/\c/\val in {
    0/0/.10,0/1/.95,0/2/.10,0/3/.90,0/4/.10,
    1/0/.10,1/1/0,1/2/.15,1/3/0,1/4/.10,
    2/0/.10,2/1/.80,2/2/.10,2/3/.75,2/4/.10,
    3/0/.10,3/1/0,3/2/.10,3/3/0,3/4/.10,
    4/0/.10,4/1/0,4/2/.10,4/3/0,4/4/.10
} {
    \pgfmathtruncatemacro{\tone}{22+68*\val}
    \filldraw[cell, fill=currentblue!\tone]
        ($(SA)+({0.42*\c},{0.42*(4-\r)})$) rectangle ++(0.42,0.42);
}
\draw[black!60, line width=0.7pt] (SA) rectangle ++(2.10,2.10);
\draw[selectorgreen, line width=1.15pt, rounded corners=1pt]
    ($(SA)+(0.44,0.02)$) rectangle ++(0.38,2.06);
\draw[selectorgreen, line width=1.15pt, rounded corners=1pt]
    ($(SA)+(1.28,0.02)$) rectangle ++(0.38,2.06);
\node[font=\small] at ($(SA)+(1.05,2.42)$) {$\mat{S}_t$};
\node[label] at ($(SA)+(1.05,-0.28)$) {value features};
\foreach \r in {1,...,5} {
    \node[font=\tiny, text=black!55, anchor=east]
        at ($(SA)+(-0.10,{2.31-0.42*\r})$) {\r};
}

\coordinate (EA) at (4.18,1.40);
\foreach \c/\val in {0/0,1/.80,2/0,3/.60,4/0} {
    \pgfmathtruncatemacro{\tone}{22+68*\val}
    \filldraw[cell, fill=selectorgreen!\tone]
        ($(EA)+(0,{0.42*(4-\c)})$) rectangle ++(0.42,0.42);
}
\draw[black!60, line width=0.7pt] (EA) rectangle ++(0.42,2.10);
\node[font=\small] at ($(EA)+(0.21,2.42)$) {$\vct{e}_t$};
\node[label] at ($(EA)+(0.21,-0.28)$) {selector};
\node[font=\large, text=black!72] at (3.70,2.48) {$\times$};

\coordinate (RA) at (6.38,1.40);
\foreach \r/\val in {0/\curzero,1/\curone,2/\curtwo,3/\curthree,4/\curfour} {
    \pgfmathtruncatemacro{\tone}{22+68*\val}
    \filldraw[cell, fill=currentblue!\tone]
        ($(RA)+(0,{0.42*(4-\r)})$) rectangle ++(0.42,0.42);
}
\draw[black!60, line width=0.7pt] (RA) rectangle ++(0.42,2.10);
\node[font=\small] at ($(RA)+(0.21,2.42)$) {$\mat{S}_t\vct{e}_t$};
\node[label, text width=1.65cm] at ($(RA)+(0.21,-0.35)$)
    {addresses 1 and 3};
\draw[flow, draw=currentblue]
    ($(EA)+(0.62,1.05)$) -- node[below, font=\tiny, text=black!68]
    {} ($(RA)+(-0.20,1.05)$);

\begin{scope}[xshift=-0.75cm]
\node[panel] (panelB) at (12.55,2.55) {};
\node[paneltitle, anchor=north] at ([yshift=-0.20cm]panelB.north)
    {(b) Value-side delta rule};

\coordinate (TB) at (8.83,3.42);
\foreach \i/\val in {0/\tarzero,1/\tarone,2/\tartwo,3/\tarthree,4/\tarfour} {
    \pgfmathtruncatemacro{\tone}{22+68*\val}
    \filldraw[cell, fill=targetorange!\tone]
        ($(TB)+(0.38*\i,0)$) rectangle ++(0.38,0.43);
}
\draw[black!55, line width=0.65pt] (TB) rectangle ++(1.90,0.43);
\node[toplabel] at ($(TB)+(0.95,0.67)$) {target: address 5};
\node[mathlabel, text=targetorange!85!black] at ($(TB)+(0.95,-0.33)$)
    {$(\vct{v}_t^\top \vct{e}_t)\vct{k}_t$};

\coordinate (RB) at (11.65,3.42);
\foreach \i/\val in {0/\curzero,1/\curone,2/\curtwo,3/\curthree,4/\curfour} {
    \pgfmathtruncatemacro{\tone}{22+68*\val}
    \filldraw[cell, fill=currentblue!\tone]
        ($(RB)+(0.38*\i,0)$) rectangle ++(0.38,0.43);
}
\draw[black!55, line width=0.65pt] (RB) rectangle ++(1.90,0.43);
\node[toplabel] at ($(RB)+(0.95,0.67)$) {read: addresses 1 and 3};
\node[mathlabel, text=currentblue!80!black] at ($(RB)+(0.95,-0.33)$)
    {$\mat{S}_t\vct{e}_t$};

\coordinate (CB) at (14.47,3.42);
\foreach \i/\val in {0/\czero,1/\cone,2/\ctwo,3/\cthree,4/\cfour} {
    \pgfmathtruncatemacro{\tone}{20+70*abs(\val)}
    \pgfmathtruncatemacro{\signcode}{\val>0.001 ? 1 : (\val<-0.001 ? -1 : 0)}
    \ifnum\signcode=1
        \filldraw[cell, fill=residualpurple!\tone]
            ($(CB)+(0.38*\i,0)$) rectangle ++(0.38,0.43);
        \node[font=\tiny, text=white] at ($(CB)+(0.38*\i+0.19,0.215)$) {$+$};
    \else
        \ifnum\signcode=-1
            \filldraw[cell, fill=currentblue!\tone]
                ($(CB)+(0.38*\i,0)$) rectangle ++(0.38,0.43);
            \node[font=\tiny, text=white] at ($(CB)+(0.38*\i+0.19,0.215)$) {$-$};
        \else
            \filldraw[cell, fill=stategray!30]
                ($(CB)+(0.38*\i,0)$) rectangle ++(0.38,0.43);
        \fi
    \fi
}
\draw[black!55, line width=0.65pt] (CB) rectangle ++(1.90,0.43);
\node[toplabel] at ($(CB)+(0.95,0.67)$) {address residual};
\node[mathlabel, text=residualpurple!85!black] at ($(CB)+(0.95,-0.33)$)
    {$\vct{c}_t$};
\node[font=\large, text=black!60] at (11.20,3.64) {$-$};
\node[font=\large, text=black!60] at (14.02,3.64) {$=$};

\coordinate (SB0) at (9.55,0.88);
\foreach \r/\val in {0/\curzero,1/\curone,2/\curtwo,3/\curthree,4/\curfour} {
    \pgfmathtruncatemacro{\tone}{22+60*\val}
    \foreach \c in {0,...,4} {
        \filldraw[cell, fill=currentblue!22]
            ($(SB0)+({0.34*\c},{0.34*(4-\r)})$) rectangle ++(0.34,0.34);
    }
    \filldraw[cell, fill=currentblue!\tone]
        ($(SB0)+(0.34,{0.34*(4-\r)})$) rectangle ++(0.34,0.34);
    \filldraw[cell, fill=currentblue!\tone]
        ($(SB0)+(1.02,{0.34*(4-\r)})$) rectangle ++(0.34,0.34);
}
\draw[black!60, line width=0.7pt] (SB0) rectangle ++(1.70,1.70);
\draw[selectorgreen, line width=0.9pt, rounded corners=0.8pt]
    ($(SB0)+(0.36,0.02)$) rectangle ++(0.30,1.66);
\draw[selectorgreen, line width=0.9pt, rounded corners=0.8pt]
    ($(SB0)+(1.04,0.02)$) rectangle ++(0.30,1.66);
\node[font=\small] at ($(SB0)+(0.85,2.00)$) {$\mat{S}_t$};
\node[tinylabel] at ($(SB0)+(0.85,-0.28)$)
    {before: addresses 1 and 3};

\coordinate (SB1) at (13.88,0.88);
\foreach \r/\val in {0/\newzero,1/\newone,2/\newtwo,3/\newthree,4/\newfour} {
    \pgfmathtruncatemacro{\tone}{22+60*\val}
    \foreach \c in {0,...,4} {
        \filldraw[cell, fill=currentblue!22]
            ($(SB1)+({0.34*\c},{0.34*(4-\r)})$) rectangle ++(0.34,0.34);
    }
    \filldraw[cell, fill=currentblue!\tone]
        ($(SB1)+(0.34,{0.34*(4-\r)})$) rectangle ++(0.34,0.34);
    \filldraw[cell, fill=currentblue!\tone]
        ($(SB1)+(1.02,{0.34*(4-\r)})$) rectangle ++(0.34,0.34);
}
\draw[black!60, line width=0.7pt] (SB1) rectangle ++(1.70,1.70);
\draw[selectorgreen, line width=0.9pt, rounded corners=0.8pt]
    ($(SB1)+(0.36,0.02)$) rectangle ++(0.30,1.66);
\draw[selectorgreen, line width=0.9pt, rounded corners=0.8pt]
    ($(SB1)+(1.04,0.02)$) rectangle ++(0.30,1.66);
\node[font=\small] at ($(SB1)+(0.85,2.00)$) {$\mat{S}_t^{\prime}$};
\node[tinylabel] at ($(SB1)+(0.85,-0.28)$)
    {after: address 5};

\draw[flow, draw=residualpurple]
    ($(SB0)+(2.05,0.84)$) --
    ($(SB1)+(-0.37,0.84)$);
\end{scope}

\end{tikzpicture}%
        \endgroup
    }
    \caption{Illustration of the value-side read and delta rule. (a) The
    selector $\vct{e}_t$ extracts the address response $\mat{S}_t\vct{e}_t$ of selected value
    features, which occupy addresses 1 and 3 in this example. (b) Comparing
    this response with the target $(\vct{v}_t^\top \vct{e}_t)\vct{k}_t$ produces a negative correction
    at addresses 1 and 3 and a positive correction at address 5, reorganizing
    the selected features toward the target address.}
    \label{fig:ddn-mechanism}
\end{figure*}
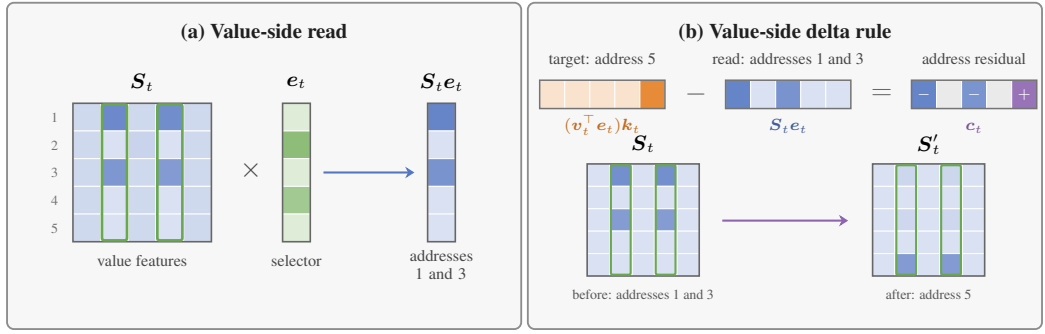

Linear attention constructs an associative state
$\mat{S}_t\in\mathbb{R}^{d_k\times d_v}$ by accumulating an outer-product update
$\vct{k}_t\vct{v}_t^\top$ from each token, where $\vct{k}_t\in\mathbb{R}^{d_k}$ and
$\vct{v}_t\in\mathbb{R}^{d_v}$. Viewed row-wise, this state $\mat{S}_t$ forms a bank of $d_k$
addresses, each carrying a $d_v$-dimensional value representation. In this
view, $\vct{k}_t$ serves two roles: it is the semantic key matched by subsequent
queries, and it acts as the write-address vector in the outer-product update
$\vct{k}_t\vct{v}_t^\top$, determining how $\vct{v}_t$ is distributed across the address bank.
For vanilla linear attention with $\mat{S}_0=0$, these two roles appear in
the state construction and readout:
\begin{alignat*}{2}
    \mat{S}_t
    &= \sum_{j=1}^{t}\vct{k}_j\vct{v}_j^\top
    &\qquad& \text{(write address)}, \\
    \mat{S}_t^\top\vct{q}_t
    &= \sum_{j=1}^{t}\vct{v}_j(\vct{k}_j^\top\vct{q}_t)
    &\qquad& \text{(semantic matching)}.
\end{alignat*}
Existing delta rules use the key-side read $\mat{S}_{t-1}^\top \vct{k}_t$ to correct the
value associated with the current key, but do not explicitly examine how
related value features are distributed across different addresses.
Building on DART's observation that value-side reads from recurrent states
can decode token-conditioned keys~\citep{qian2026dart}, we exploit this
complementary operation to guide associative memory updates.
While linear attention is designed for key-to-value retrieval and a value-side
read does not generally invert this mapping to recover semantic keys, this
read can reveal how selected value features are represented across the
recurrent state's addresses.
Specifically, given a value-side selector $\vct{e}_t\in\mathbb{R}^{d_v}$,
the response $\mat{S}_{t-1}\vct{e}_t\in\mathbb{R}^{d_k}$ aggregates the
address directions associated with the selected value features, as illustrated
in Figure~\ref{fig:ddn-mechanism}(a).
The resulting state-dependent address vector can guide the reorganization of
related value content within the associative memory.

Motivated by this observation, we propose Dual DeltaNet (DDN),
which extends KDA with complementary key-side and value-side delta rules.
While the key-side delta rule corrects the value associated with the current key $\vct{k}_t$,
the value-side delta rule adjusts the addresses of value features.
Specifically, DDN constructs a selector $\vct{e}_t$ by applying a sigmoid channel gate to
the normalized value direction $\hat{\vct{v}}_t$.
The value-side read $\mat{S}_t \vct{e}_t$ aggregates stored address directions weighted by
their alignment with the selected value features.
For each outer-product write $\vct{k}_j\vct{v}_j^\top$, the value-side read gives
$(\vct{k}_j\vct{v}_j^\top)\vct{e}_t=(\vct{v}_j^\top \vct{e}_t)\vct{k}_j$, weighting its write address by the
selected value features. The value-side delta rule moves the aggregated
address toward $(\vct{v}_t^\top \vct{e}_t)\vct{k}_t$, the corresponding address
of the current key--value association, as illustrated in Figure~\ref{fig:ddn-mechanism}(b).
This correction reorganizes value-conditioned address occupancy within the state,
potentially reducing redundant storage and interference.
Our contributions are summarized as follows:
\begin{itemize}
    \item We introduce Dual DeltaNet (DDN), which applies a key-side delta rule
    to correct the value associated with the current key and a complementary
    value-side delta rule to reorganize value content across addresses
    within the same associative state.
    \item We develop efficient training and inference algorithms for DDN.
    For training, we extend KDA's chunkwise parallel formulation using two-sided compact WY factorizations.
    For inference, we express the recurrence in a rank-2 factorized form that increases theoretical arithmetic intensity during memory-bound decoding.
    Our experiments show modest
    training and inference overhead relative to KDA, with efficiency
    competitive with GDN-2 and Mamba-3 MIMO.
    \item We compare DDN with KDA, GDN-2, and Mamba-3 on language-model
    pretraining, zero-shot downstream tasks, real-world retrieval tasks, and
    synthetic NIAH tasks. DDN achieves the strongest overall performance
    across the 130M, 340M, and 1B scales, with the most pronounced gains on
    the needle-in-a-haystack benchmark.
\end{itemize}

\section{Background}

\paragraph{Linear Attention.}
Linear attention admits a recurrence
that accumulates key--value information through outer-product updates
\citep{katharopoulos2020transformers}. Let
$\vct{q}_t,\vct{k}_t\in\mathbb{R}^{d_k}$ and $\vct{v}_t\in\mathbb{R}^{d_v}$. The recurrent
state $\mat{S}_t\in\mathbb{R}^{d_k\times d_v}$ and its readout are
\begin{align}
    \mat{S}_t &= \mat{S}_{t-1}+\vct{k}_t\vct{v}_t^\top,
    &
    \vct{o}_t &= \mat{S}_t^\top \vct{q}_t.
    \label{eq:background_linear_attention}
\end{align}
For $\mat{S}_0=0$, expanding the recurrence gives
\begin{align}
    \vct{o}_t=\sum_{j=1}^{t}\vct{v}_j\bigl(\vct{k}_j^\top \vct{q}_t\bigr),
    \label{eq:background_linear_attention_sum}
\end{align}
so linear attention combines the values $\vct{v}_j$ using the key--query inner
products $\vct{k}_j^\top \vct{q}_t$ as weights.

\paragraph{Delta Rule.}
DeltaNet replaces the additive write with a key-side correction
\citep{schlag2021linear},
\begin{align}
    \mat{S}_t
    &=
    \mat{S}_{t-1}+\beta_t\vct{k}_t
    \bigl(\vct{v}_t-\mat{S}_{t-1}^\top \vct{k}_t\bigr)^\top
    \nonumber\\
    &=
    \left(\mat{I}-\beta_t\vct{k}_t\vct{k}_t^\top\right)\mat{S}_{t-1}
    +\beta_t\vct{k}_t\vct{v}_t^\top.
    \label{eq:background_delta_rule}
\end{align}
The residual $\vct{v}_t-\mat{S}_{t-1}^\top \vct{k}_t$ measures the difference between the
incoming value and the value currently associated with $\vct{k}_t$, while
$\beta_t\in(0,1)$ controls the strength of this correction.

Several subsequent models extend DeltaNet with more flexible forgetting and
update mechanisms. Gated DeltaNet adds adaptive forgetting to the delta rule
\citep{yang2025gated}. Kimi Delta Attention (KDA) further generalizes
forgetting to channel-wise decay~\citep{team2025kimi}:
\begin{equation}
    \mat{S}_t=\left(\mat{I}-\beta_t\vct{k}_t\vct{k}_t^\top\right)\mat{D}_t\mat{S}_{t-1}
    +\beta_t\vct{k}_t\vct{v}_t^\top.
    \label{eq:background_kda}
\end{equation}
where $\mat{D}_t=\operatorname{Diag}(\vct{\alpha}_t)$ and
$\vct{\alpha}_t\in(0,1)^{d_k}$ allows the value content stored at each address to
decay at a different rate. Gated DeltaNet-2 further separates channel-wise
decay, erasure, and writing
\citep{hatamizadeh2026gated}.

\section{Methodology}
\label{sec:ddn}

DDN maintains a matrix state $\mat{S}_t\in\mathbb{R}^{d_k\times d_v}$.
The query $\vct{q}_t$ and key $\vct{k}_t$ are L2-normalized in key space before
the recurrent update. Let $\beta_t\in(0,1)$ be the KDA update strength and let
$\mat{D}_t=\operatorname{Diag}(\vct{\alpha}_t)$ with $\vct{\alpha}_t\in(0,1)^{d_k}$. The retained
state after decay is $\bar{\mat{S}}_{t-1}=\mat{D}_t\mat{S}_{t-1}$.

\paragraph{Key-side delta rule.}
KDA updates the state through a value residual:
\begin{equation}
    \mat{S}_t^{\mathrm{KDA}}
    =\bar{\mat{S}}_{t-1}
    +\beta_t\vct{k}_t\left(\vct{v}_t-\bar{\mat{S}}_{t-1}^\top \vct{k}_t\right)^\top.
    \label{eq:kda-update}
\end{equation}
This form adjusts the incoming value $\vct{v}_t$ by subtracting the value already
associated with $\vct{k}_t$ in $\bar{\mat{S}}_{t-1}$. The resulting correction is written
at the same key with strength $\beta_t$.
Equivalently, the update can be expressed as erasure followed by writing:
\begin{equation}
    \mat{S}_t^{\mathrm{KDA}}
    =\left(\mat{I}-\beta_t\vct{k}_t\vct{k}_t^\top\right)\bar{\mat{S}}_{t-1}
    +\beta_t\vct{k}_t\vct{v}_t^\top.
    \label{eq:kda-erase-write}
\end{equation}
The left factor removes a fraction $\beta_t$
of the value associated with $\vct{k}_t$ before the new value is written.
Denote this factor by $\mat{A}_t=\mat{I}-\beta_t\vct{k}_t\vct{k}_t^\top$.

\paragraph{Dual DeltaNet.}
Let $\hat{\vct{v}}_t=\vct{v}_t/\lVert \vct{v}_t\rVert_2$ denote the normalized value direction.
To select which value features
guide the correction, define the channel gate
$\vct{m}_t=\operatorname{sigmoid}(\vct{s}_t^m)\in(0,1)^{d_v}$, where the logit vector
$\vct{s}_t^m\in\mathbb R^{d_v}$ is predicted from the current input. Define the
normalized selector and its target scale as
\begin{align}
    \vct{e}_t=\frac{\vct{m}_t\odot\hat{\vct{v}}_t}
    {\lVert \vct{m}_t\odot\hat{\vct{v}}_t\rVert_2},
    \quad
    \tau_t=\vct{v}_t^\top \vct{e}_t.
    \label{eq:ddn-gated-probe}
\end{align}
Let $\gamma_t=\operatorname{sigmoid}(s_t^\gamma)$ be a scalar gate whose
logit $s_t^\gamma$ is predicted from the current input. The complementary value-side delta rule then applies
\begin{equation}
    \mat{S}_t=\mat{S}_t^{\mathrm{KDA}}+\gamma_t
    \left(\tau_t\vct{k}_t-\mat{S}_t^{\mathrm{KDA}}\vct{e}_t\right)\hat{\vct{v}}_t^\top.
    \label{eq:ddn-address}
\end{equation}
The channel gate selects the value features used by
$\mat{S}_t^{\mathrm{KDA}}\vct{e}_t$ to identify matching addresses.
The residual guides the correction of the resulting address vector toward
$\tau_t\vct{k}_t$, with strength controlled by $\gamma_t$.
Equivalently, the update can be written as:
\begin{align}
    \mat{S}_t
    &=\mat{S}_t^{\mathrm{KDA}}\mat{B}_t+\gamma_t\tau_t\vct{k}_t\hat{\vct{v}}_t^\top \notag\\
    &=\mat{A}_t\bar{\mat{S}}_{t-1}\mat{B}_t+\eta_t\vct{k}_t\hat{\vct{v}}_t^\top,
    \label{eq:ddn-delta}
\end{align}
where $\mat{B}_t=\mat{I}-\gamma_t\vct{e}_t\hat{\vct{v}}_t^\top$.
Since $\vct{v}_t=\lVert \vct{v}_t\rVert_2\hat{\vct{v}}_t$, the effective write strength is
\begin{equation}
    \eta_t=\beta_t\lVert \vct{v}_t\rVert_2
    +\gamma_t(1-\beta_t)\tau_t.
    \label{eq:ddn-effective-write-strength}
\end{equation}

\paragraph{Rank-2 factorized form.}
Autoregressive decoding with linear attention is often memory-bound because
each token reads and writes the recurrent state while performing relatively
little computation. Mamba-3's MIMO formulation improves model expressivity by
increasing computation while keeping recurrent-state memory traffic largely
unchanged~\citep{lahoti2026mamba3}. This increases arithmetic intensity (FLOPs per input/output byte)
and improves hardware utilization during decoding.
Motivated by this principle, DDN expresses its two rank-one corrections as a
rank-2 update in Algorithm~\ref{alg:ddn-rank-two}.
Both state-dependent reads are computed from the decayed state $\bar{\mat{S}}_{t-1}$,
and the resulting correction directions are packed on chip
into $\mathcal K_t\in\mathbb R^{d_k\times2}$ and $\mathcal V_t\in\mathbb R^{2\times d_v}$.
Table~\ref{tab:decode-arithmetic-intensity} compares the input/output traffic,
FLOPs, and arithmetic intensity of KDA and DDN. Both methods load one
$d_k\times d_v$ state and write it back once per token, while DDN increases the leading-order
computation from $7d_kd_v$ to $11d_kd_v$ with only
$\mathcal O(d_v)$ additional input. Its arithmetic intensity consequently
increases from approximately $1.75$ to $2.75$ FLOPs/byte when state traffic dominates.

\begin{algorithm}[htbp]
\caption{Rank-2 factorized form}
\label{alg:ddn-rank-two}
\footnotesize
\setlength{\abovedisplayskip}{2pt}
\setlength{\belowdisplayskip}{2pt}
\setlength{\abovedisplayshortskip}{2pt}
\setlength{\belowdisplayshortskip}{2pt}
\begin{algorithmic}[1]
\REQUIRE $\mat{S}_{t-1}$, $\mat{D}_t$, $\vct{k}_t$, $\vct{v}_t$, $\vct{e}_t$, $\vct{q}_t$, $\beta_t$, and
$\gamma_t$
\ENSURE Updated state $\mat{S}_t$ and output $\vct{o}_t$
\STATE \textbf{Preparation:}\quad
Compute $\bar{\mat{S}}_{t-1}$, $\hat{\vct{v}}_t$, and $\tau_t$.
\STATE \textbf{Read both sides:}\quad
$\vct{\ell}_t\leftarrow\bar{\mat{S}}_{t-1}^\top \vct{k}_t$, \quad
$\vct{r}_t\leftarrow\bar{\mat{S}}_{t-1}\vct{e}_t$.
\STATE \textbf{Form residual directions.}
\begin{equation*}
    \begin{aligned}
        \tilde{\vct{v}}_t&\leftarrow \vct{v}_t-\vct{\ell}_t,\\
        \tilde{\vct{k}}_t&\leftarrow
        (\tau_t-\beta_t\tilde{\vct{v}}_t^\top \vct{e}_t)\vct{k}_t-\vct{r}_t.
    \end{aligned}
\end{equation*}
\STATE \textbf{Pack rank-2 factors.}
\begin{equation*}
    \mathcal K_t\leftarrow
    \begin{bmatrix}\vct{k}_t&\tilde{\vct{k}}_t\end{bmatrix},
    \qquad
    \mathcal V_t\leftarrow
    \begin{bmatrix}
        \beta_t\tilde{\vct{v}}_t^\top\\
        \gamma_t\hat{\vct{v}}_t^\top
    \end{bmatrix}.
\end{equation*}
\STATE \textbf{Update state:}\quad
$\mat{S}_t\leftarrow\bar{\mat{S}}_{t-1}+\mathcal K_t\mathcal V_t$.
\STATE \textbf{Read output:}\quad
$\vct{o}_t\leftarrow d_k^{-1/2}\mat{S}_t^\top \vct{q}_t$.
\end{algorithmic}
\end{algorithm}

\begin{table}[htbp]
\caption{Arithmetic intensity for (a) KDA and (b) DDN, measured in
FLOPs/byte for 2-byte inputs and outputs.
The batch and head dimensions cancel out.
DDN's higher arithmetic intensity can improve hardware utilization during
memory-bound phases such as autoregressive decoding.}
\label{tab:decode-arithmetic-intensity}
\centering
\renewcommand{\arraystretch}{1.25}
\resizebox{\linewidth}{!}{%
\begin{tabular}{@{}llll@{}}
\toprule
Input & Output & FLOPs & Arithmetic intensity \\
\midrule
$\begin{gathered}
\mat{S}_{t-1}:(d_k,d_v)\\
\vct{\alpha}_t,\vct{k}_t,\vct{q}_t:(d_k)\\
\vct{v}_t:(d_v)\\
\beta_t:(1)
\end{gathered}$
& $\begin{gathered}
\mat{S}_t:(d_k,d_v)\\
\vct{o}_t:(d_v)
\end{gathered}$
& $7d_kd_v+\mathcal O(d_k+d_v)$
& $\displaystyle
\frac{7d_kd_v+\mathcal O(d_k+d_v)}
{2(2d_kd_v+3d_k+2d_v+1)}
\approx 1.75$ \\
\bottomrule
\end{tabular}%
}
\vspace{2pt}

\textnormal{(a) KDA (2-byte data).}
\vspace{6pt}

\resizebox{\linewidth}{!}{%
\begin{tabular}{@{}llll@{}}
\toprule
Input & Output & FLOPs & Arithmetic intensity \\
\midrule
$\begin{gathered}
\mat{S}_{t-1}:(d_k,d_v)\\
\vct{\alpha}_t,\vct{k}_t,\vct{q}_t:(d_k)\\
\vct{v}_t,\vct{e}_t:(d_v)\\
\beta_t,\gamma_t:(1)
\end{gathered}$
& $\begin{gathered}
\mat{S}_t:(d_k,d_v)\\
\vct{o}_t:(d_v)
\end{gathered}$
& $11d_kd_v+\mathcal O(d_k+d_v)$
& $\displaystyle
\frac{11d_kd_v+\mathcal O(d_k+d_v)}
{2(2d_kd_v+3d_k+3d_v+2)}
\approx 2.75$ \\
\bottomrule
\end{tabular}%
}
\vspace{2pt}

\textnormal{(b) DDN (2-byte data).}
\end{table}

\begin{figure}[!ht]
\centering
\definecolor{fgate_color}{RGB}{252,224,225}
\definecolor{selector_color}{RGB}{255,210,210}
\definecolor{delta_color}{RGB}{242,243,193}
\definecolor{add_norm_color}{RGB}{252,226,187}
\definecolor{silu_color}{RGB}{203,231,207}
\definecolor{linear_color}{RGB}{220,223,240}
\definecolor{conv_color}{RGB}{252,224,225}
\definecolor{l2_color}{RGB}{252,226,187}
\definecolor{gray_bbox_color}{RGB}{243,243,244}
\definecolor{oproj_color}{RGB}{220,223,240}

\tikzset{
  gdelta/.style={
    draw=black,
    very thick,
    fill=gray_bbox_color,
    minimum width=300pt,
    minimum height=235pt,
    rounded corners=10pt
  },
  norm/.style={
    draw=black,
    very thick,
    fill=add_norm_color!80,
    minimum width=44pt,
    rounded corners=3pt,
    align=center
  },
  stacked/.style={
    draw=black,
    very thick,
    fill=linear_color!80,
    minimum width=38pt,
    minimum height=15pt,
    rounded corners=3pt
  },
  conv/.style={
    draw=black,
    very thick,
    minimum width=32pt,
    fill=conv_color!80,
    rounded corners=3pt,
    rectangle,
    font=\small
  },
  ltwo/.style={
    draw=black,
    very thick,
    minimum width=32pt,
    fill=l2_color!80,
    rounded corners=3pt,
    rectangle,
    font=\small
  },
  layerlink/.style={
    -{Latex[length=4pt]},
    very thick
  },
  normlink/.style={very thick},
  residual/.style={very thick,rounded corners=5pt},
  qf/.style={-{Latex[length=4pt]},very thick,rounded corners=5pt},
  otimes/.style={
    draw=black,
    very thick,
    circle,
    minimum size=9pt,
    inner sep=0pt,
    outer sep=0pt,
    path picture={
      \draw (path picture bounding box.center) -- ++(0.12cm,0.12cm)
            (path picture bounding box.center) -- ++(-0.12cm,-0.12cm)
            (path picture bounding box.center) -- ++(-0.12cm,0.12cm)
            (path picture bounding box.center) -- ++(0.12cm,-0.12cm);
    }
  },
  sigmoid/.style={
    draw=black,
    very thick,
    circle,
    minimum size=10pt,
    inner sep=0pt,
    outer sep=0pt,
    path picture={
      \draw[domain=-2:2,samples=40,variable=\x,blue,thick]
        plot ({\x*0.13},{0.32/(1+exp(-2.5*\x))-0.16});
    }
  },
  swish/.style={
    draw=black,
    very thick,
    circle,
    minimum size=9pt,
    inner sep=0pt,
    outer sep=0pt,
    path picture={
      \draw[domain=-1.4:1.4,samples=40,variable=\x,blue,thick]
        plot ({\x*0.16},{0.34*\x/(1+exp(-2*\x))});
    }
  },
  kgdelta/.style={
    draw=black,
    very thick,
    fill=delta_color!50,
    minimum width=230pt,
    minimum height=0.9cm,
    rounded corners=3pt,
    align=center
  }
}

\begin{tikzpicture}

\node[gdelta] (gdelta) at (0,0) {};
\node[below=12pt] at (gdelta.south) (input) {Input};

\node[kgdelta,yshift=18pt] at (gdelta.center) (kernel)
  {Dual DeltaNet
   };

\node[stacked,trapezium,trapezium left angle=110,
  trapezium right angle=110,inner xsep=1pt,outer sep=0pt,
  minimum width=31pt,anchor=south,yshift=20pt]
  at ($(gdelta.south west)!0.14!(gdelta.south east)$) (qproj) {Linear};
\node[conv,anchor=south,yshift=8pt] at (qproj.north) (qconv) {Conv};
\node[swish,anchor=south,yshift=4pt] at (qconv.north) (qsilu) {};
\node[ltwo,anchor=south,yshift=4pt] at (qsilu.north) (ql2) {L2};
\draw[normlink] (qproj.north) -- (qconv.south);
\draw[normlink] (qconv.north) -- (qsilu.south);
\draw[normlink] (qsilu.north) -- (ql2.south);
\draw[layerlink] (ql2.north) -- (ql2.north|-kernel.south);

\node[stacked,trapezium,trapezium left angle=110,
  trapezium right angle=110,inner xsep=1pt,outer sep=0pt,
  minimum width=31pt,anchor=south,yshift=18pt,opacity=0.25]
  at ($(gdelta.south west)!0.18!(gdelta.south east)$) (kproj)
  {\textcolor{white}{Linear}};
\node[conv,anchor=south,yshift=8pt,opacity=0.25]
  at (kproj.north) (kconv) {\textcolor{white}{Conv}};
\node[swish,anchor=south,yshift=4pt,opacity=0.25]
  at (kconv.north) (ksilu) {};
\node[ltwo,anchor=south,yshift=4pt,opacity=0.25]
  at (ksilu.north) (kl2) {\textcolor{white}{L2}};
\draw[normlink,opacity=0.25] (kproj.north) -- (kconv.south);
\draw[normlink,opacity=0.25] (kconv.north) -- (ksilu.south);
\draw[normlink,opacity=0.25] (ksilu.north) -- (kl2.south);
\draw[layerlink,opacity=0.25] (kl2.north) -- (kl2.north|-kernel.south);

\node[stacked,trapezium,trapezium left angle=110,
  trapezium right angle=110,inner xsep=1pt,outer sep=0pt,
  minimum width=31pt,anchor=south,yshift=20pt]
  at ($(gdelta.south west)!0.36!(gdelta.south east)$) (vproj) {Linear};
\node[conv,anchor=south,yshift=8pt] at (vproj.north) (vconv) {Conv};
\node[swish,anchor=south,yshift=4pt] at (vconv.north) (vsilu) {};
\draw[normlink] (vproj.north) -- (vconv.south);
\draw[normlink] (vconv.north) -- (vsilu.south);
\draw[layerlink] (vsilu.north) -- (vsilu.north|-kernel.south);

\node[stacked,isosceles triangle,isosceles triangle apex angle=105,
  inner sep=.8pt,anchor=south,yshift=20pt,shape border rotate=90]
  at ($(gdelta.south west)!0.53!(gdelta.south east)$) (aproj) {Lin.};
\draw[qf] ($(aproj.north)-(0,2pt)$) -- (aproj.north|-kernel.south);

\node[stacked,isosceles triangle,isosceles triangle apex angle=105,
  inner sep=.8pt,anchor=south,yshift=20pt,
  shape border rotate=90]
  at ($(gdelta.south west)!0.78!(gdelta.south east)$) (gatesproj)
  {Lin. $\times 3$};
\node[sigmoid,anchor=south,yshift=4pt] at (gatesproj.north) (gatessigmoid) {};
\draw[normlink] (gatesproj.north) -- (gatessigmoid.south);
\draw[qf] (gatessigmoid.north) -- (gatessigmoid.north|-kernel.south);

\coordinate (labelrow) at ([yshift=-10pt]kernel.south);
\node[anchor=east] at ([xshift=-5pt]qproj.north|-labelrow) {$\vct{q},\vct{k}$};
\node[anchor=west] at ([xshift=5pt]vproj.north|-labelrow) {$\vct{v}$};
\node[anchor=east] at ([xshift=-5pt]aproj.north|-labelrow) {$\vct{\alpha}$};
\node[anchor=east] at ([xshift=-5pt]gatesproj.north|-labelrow)
  {$\beta,\,\vct{m},\,\gamma$};

\draw[residual] (input.north) -- (gdelta.south) -- ++(0,10pt)
  -| (qproj.south);
\draw[residual,opacity=0.25] (input.north) -- (gdelta.south) -- ++(0,10pt)
  -| (kproj.south);
\draw[residual] (input.north) -- (gdelta.south) -- ++(0,10pt)
  -| (vproj.south);
\draw[residual] (input.north) -- (gdelta.south) -- ++(0,10pt)
  -| (aproj.south);
\draw[residual] (input.north) -- (gdelta.south) -- ++(0,10pt)
  -| (gatesproj.south);

\node[norm,anchor=south,yshift=10pt] at (kernel.north) (outnorm) {Norm};
\node[otimes,anchor=south,yshift=6pt] at (outnorm.north) (otimes) {};
\node[draw=black,very thick,fill=oproj_color!80,minimum width=78pt,
  minimum height=10pt,shape=trapezium,trapezium angle=45,
  rounded corners=3pt,anchor=south,yshift=6pt]
  at (otimes.north) (oproj) {Linear};
\draw[layerlink] (kernel.north) -- (outnorm.south);
\draw[normlink] (outnorm.north) -- (otimes.south);
\draw[normlink] (otimes.north) -- (oproj.south);

\coordinate (outgatex) at
  ($(gdelta.south west)!0.90!(gdelta.south east)$);
\node[stacked,trapezium,trapezium left angle=110,
  trapezium right angle=110,inner xsep=1pt,outer sep=0pt,
  minimum width=32pt,anchor=center]
  at (outgatex |- outnorm.center)
  (outgateproj) {Linear};
\node[swish] at (outgateproj.north |- otimes.center) (outgatesilu) {};
\draw[normlink] (outgateproj.north) -- (outgatesilu.south);
\draw[residual] (input.north) -- (gdelta.south)
  |- ([xshift=10pt,yshift=10pt]gdelta.south) -| (outgateproj.south);
\draw[normlink] (outgatesilu.west) -- (otimes.east);

\node[above=12pt] at (gdelta.north) (output) {Output};
\draw[layerlink] (oproj.north) -- (output.south);

\end{tikzpicture}
\caption{DDN block design. Query, key, and value branches apply linear
projections, short causal convolutions, and SiLU activations, with
$\ell_2$ normalization applied to queries and keys. The control branches
produce the channel-wise decay $\vct{\alpha}_t$ and the sigmoid-gated quantities
$\beta_t$, $\vct{m}_t$, and $\gamma_t$.}
\label{fig:ddn-block}
\end{figure}
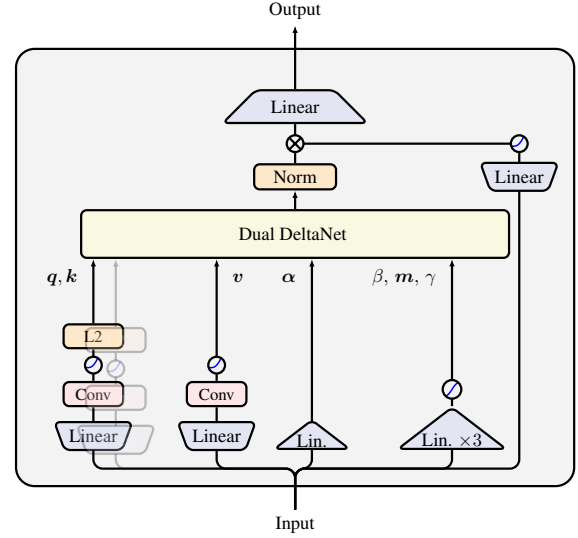

\paragraph{Block design.}
DDN follows KDA's block design, as illustrated in
Figure~\ref{fig:ddn-block}. The query, key, and value branches each apply a
linear projection, a short causal convolution, and a SiLU activation, with
additional $\ell_2$ normalization for queries and keys. The gates $\beta_t$,
$\vct{m}_t$, and $\gamma_t$ are produced by linear projections of the current
input followed by sigmoid activations. The decay branch follows KDA's
parameterization for $\vct{\alpha}_t$, and its nonlinear transformations are
omitted from the figure for clarity and detailed in
Appendix~\ref{app:implementation-details}.

\section{Chunkwise Computation}
\label{sec:chunk-computation}

\paragraph{Chunk notation.}
For chunks of length $C$, indexed by $n=0,1,\ldots$, write
$x_{[n]}^{i}=x_{nC+i}$ for position $i\in\{1,\ldots,C\}$.
The incoming and outgoing states are
\begin{equation}
    \mat{S}_{[n]}:=\mat{S}_{[n]}^{0}=\mat{S}_{nC},
    \qquad \mat{S}_{[n+1]}=\mat{S}_{[n]}^{C}.
    \label{eq:chunk-state-indexing}
\end{equation}
Chunk matrices carry the same subscript $[n]$.

\paragraph{Equivalent decay-free recurrence.}
Define the cumulative diagonal decay as
\begin{align}
    \vct{\pi}_{[n]}^{i}
    &=\vct{\alpha}_{[n]}^{i}\odot\vct{\alpha}_{[n]}^{i-1}\odot\cdots\odot\vct{\alpha}_{[n]}^{1}, \notag\\
    \mat{\Pi}_{[n]}^{i}&=\operatorname{Diag}(\vct{\pi}_{[n]}^{i}),
    \label{eq:chunk-cumulative-decay}
\end{align}
with $\mat{\Pi}_{[n]}^{0}=\mat{I}$.
Set $\widehat{\mat{S}}_{[n]}^{i}=(\mat{\Pi}_{[n]}^{i})^{-1}\mat{S}_{[n]}^{i}$,
$\bar{\vct{k}}_{[n]}^{i}=(\mat{\Pi}_{[n]}^{i})^{-1}\vct{k}_{[n]}^{i}$, and
$\bar{\vct{u}}_{[n]}^{i}=\beta_{[n]}^{i}\mat{\Pi}_{[n]}^{i}\vct{k}_{[n]}^{i}$.
Substituting $\mat{S}_{[n]}^{i-1}=\mat{\Pi}_{[n]}^{i-1}\widehat{\mat{S}}_{[n]}^{i-1}$
into Equation~\eqref{eq:ddn-delta}, left-multiplying by
$(\mat{\Pi}_{[n]}^{i})^{-1}$, and using
$\mat{D}_{[n]}^{i}\mat{\Pi}_{[n]}^{i-1}=\mat{\Pi}_{[n]}^{i}$ gives
\begin{align}
    \widehat{\mat{S}}_{[n]}^{i}
    =\bar{\mat{A}}_{[n]}^{i}\widehat{\mat{S}}_{[n]}^{i-1}\mat{B}_{[n]}^{i}
    +\eta_{[n]}^{i}\bar{\vct{k}}_{[n]}^{i}(\hat{\vct{v}}_{[n]}^{i})^\top,
    \label{eq:chunk-decay-free}
\end{align}
where the normalized left-side erasure is
\begin{align}
    \bar{\mat{A}}_{[n]}^{i}
    =(\mat{\Pi}_{[n]}^{i})^{-1}\mat{A}_{[n]}^{i}\mat{\Pi}_{[n]}^{i}
    =\mat{I}-\bar{\vct{k}}_{[n]}^{i}(\bar{\vct{u}}_{[n]}^{i})^\top.
    \label{eq:chunk-normalized-left-factor}
\end{align}
The right-side transition from Equation~\eqref{eq:ddn-delta} is
$\mat{B}_{[n]}^{i}=\mat{I}-\gamma_{[n]}^{i}\vct{e}_{[n]}^{i}
(\hat{\vct{v}}_{[n]}^{i})^\top$.
The initial normalized state is $\widehat{\mat{S}}_{[n]}^{0}=\mat{S}_{[n]}$.

\paragraph{Compact WY representation.}
Unrolling separates the transformed initial state from the propagated writes:
\begin{align}
    \widehat{\mat{S}}_{[n]}^{C}
    =\mat{P}_{L,[n]}\mat{S}_{[n]}\mat{P}_{R,[n]}
    +\sum_{i=1}^{C}\eta_{[n]}^{i}
    \vct{h}_{[n]}^{i}(\vct{z}_{[n]}^{i})^\top,
    \label{eq:chunk-unrolled}
\end{align}
where $\mat{P}_{L,[n]}=\bar{\mat{A}}_{[n]}^{C}\cdots \bar{\mat{A}}_{[n]}^{1}$,
$\mat{P}_{R,[n]}=\mat{B}_{[n]}^{1}\cdots \mat{B}_{[n]}^{C}$, and
\begin{align}
    \vct{h}_{[n]}^{i}
    &=(\bar{\mat{A}}_{[n]}^{C}\cdots \bar{\mat{A}}_{[n]}^{i+1})\bar{\vct{k}}_{[n]}^{i}, \notag\\
    (\vct{z}_{[n]}^{i})^\top
    &=(\hat{\vct{v}}_{[n]}^{i})^\top(\mat{B}_{[n]}^{i+1}\cdots \mat{B}_{[n]}^{C}).
    \label{eq:chunk-effective-directions}
\end{align}
Empty products are identities.

Stack $\vct{h}_{[n]}^{i}$ as columns of $\mat{H}_{[n]}$, and
$(\vct{z}_{[n]}^{i})^\top$, $(\bar{\vct{k}}_{[n]}^{i})^\top$,
$(\bar{\vct{u}}_{[n]}^{i})^\top$, $(\vct{e}_{[n]}^{i})^\top$, and
$(\hat{\vct{v}}_{[n]}^{i})^\top$ as rows of
$\mat{Z}_{[n]}$, $\bar{\mat{K}}_{[n]}$, $\bar{\mat{U}}_{[n]}$, $\mat{E}_{[n]}$, and
$\hat{\mat{V}}_{[n]}$, respectively. The compact WY representation of the left and right
transition products (Lemma~\ref{lem:compact-wy},
Appendix~\ref{app:compact-wy}) gives
\begin{align}
    \mat{H}_{[n]}(\mat{I}+\mat{T}_{L,[n]})&=\bar{\mat{K}}_{[n]}^\top, \notag\\
    (\mat{I}+\mat{T}_{R,[n]})^\top \mat{Z}_{[n]}&=\hat{\mat{V}}_{[n]},
    \label{eq:chunk-triangular-factor}
\end{align}
where
\begin{align}
  \mat{T}_{L,[n]}&=\operatorname{tril}(\bar{\mat{U}}_{[n]}\bar{\mat{K}}_{[n]}^\top,-1),\notag\\
  \mat{T}_{R,[n]}&=\operatorname{tril}
  (\mat{\Gamma}_{[n]}\mat{E}_{[n]}\hat{\mat{V}}_{[n]}^\top,-1),\notag\\
  \mat{\Gamma}_{[n]}&=\operatorname{Diag}
  (\gamma_{[n]}^{1},\ldots,\gamma_{[n]}^{C}).
  \label{eq:chunk-erasure-coefficients}
\end{align}
Writing $\mat{M}_{L,[n]}=(\mat{I}+\mat{T}_{L,[n]})^{-1}$ and
$\mat{M}_{R,[n]}=(\mat{I}+\mat{T}_{R,[n]})^{-1}$, the solutions are
\begin{equation}
    \mat{H}_{[n]}=\bar{\mat{K}}_{[n]}^\top \mat{M}_{L,[n]},
    \qquad \mat{Z}_{[n]}=\mat{M}_{R,[n]}^\top\hat{\mat{V}}_{[n]}.
    \label{eq:chunk-propagated-factors}
\end{equation}

\paragraph{State transition.}
Applying the product identity in Lemma~\ref{lem:compact-wy} to both
transition products gives
\begin{align}
    \mat{P}_{L,[n]}
    &=\mat{I}-\sum_{i=1}^{C}\vct{h}_{[n]}^{i}(\bar{\vct{u}}_{[n]}^{i})^\top
    =\mat{I}-\mat{H}_{[n]}\bar{\mat{U}}_{[n]}, \notag\\
    \mat{P}_{R,[n]}
    &=\mat{I}-\sum_{i=1}^{C}\gamma_{[n]}^{i}
    \vct{e}_{[n]}^{i}(\vct{z}_{[n]}^{i})^\top \notag\\
    &=\mat{I}-\mat{E}_{[n]}^\top\mat{\Gamma}_{[n]}\mat{Z}_{[n]}.
    \label{eq:chunk-transition-products}
\end{align}
Define the diagonal write-strength matrix
\begin{equation}
    \mat{N}_{[n]}=\operatorname{Diag}
    (\eta_{[n]}^{1},\ldots,\eta_{[n]}^{C}).
    \label{eq:chunk-write-strength}
\end{equation}
Substituting into Equation~\eqref{eq:chunk-unrolled} yields
\begin{align}
    \widehat{\mat{S}}_{[n]}^{C}
    &=(\mat{I}-\mat{H}_{[n]}\bar{\mat{U}}_{[n]})\mat{S}_{[n]}
    (\mat{I}-\mat{E}_{[n]}^\top\mat{\Gamma}_{[n]}\mat{Z}_{[n]}) \notag\\
    &\quad+\mat{H}_{[n]}\mat{N}_{[n]}\mat{Z}_{[n]},
    \label{eq:chunk-boundary-normalized}
\end{align}
Using $\mat{Z}_{[n]}=\mat{M}_{R,[n]}^\top\hat{\mat{V}}_{[n]}$ from
Equation~\eqref{eq:chunk-propagated-factors}, the right-side erasure term
factors as $\mat{E}_{[n]}^\top\mat{\Gamma}_{[n]}\mat{Z}_{[n]}
=\left(\mat{M}_{R,[n]}\mat{\Gamma}_{[n]}\mat{E}_{[n]}\right)^\top\hat{\mat{V}}_{[n]}$.
Define the right-transition factor and state-dependent quantities
\begin{align}
    \mat{F}_{R,[n]}&=\mat{M}_{R,[n]}\mat{\Gamma}_{[n]}\mat{E}_{[n]}, \notag\\
    \mat{R}_{[n]}&=\mat{S}_{[n]}\mat{P}_{R,[n]}
    =\mat{S}_{[n]}-\mat{S}_{[n]}\mat{F}_{R,[n]}^\top\hat{\mat{V}}_{[n]}, \notag\\
    \mat{\Psi}_{[n]}&=\mat{N}_{[n]}\mat{Z}_{[n]}-\bar{\mat{U}}_{[n]}\mat{R}_{[n]}.
    \label{eq:chunk-state-dependent-factors}
\end{align}
Substituting
Equation~\eqref{eq:chunk-state-dependent-factors} into
Equation~\eqref{eq:chunk-boundary-normalized} gives
\begin{align}
    \widehat{\mat{S}}_{[n]}^{C}
    &=(\mat{I}-\mat{H}_{[n]}\bar{\mat{U}}_{[n]})\mat{R}_{[n]}
    +\mat{H}_{[n]}\mat{N}_{[n]}\mat{Z}_{[n]} \notag\\
    &=\mat{R}_{[n]}+\mat{H}_{[n]}\mat{\Psi}_{[n]}.
    \label{eq:chunk-boundary-collected}
\end{align}
Restoring the cumulative decay gives the boundary recurrence
\begin{equation}
    \mat{S}_{[n+1]}=\mat{\Pi}_{[n]}^{C}
    \left(\mat{R}_{[n]}+\mat{H}_{[n]}\mat{\Psi}_{[n]}\right).
    \label{eq:chunk-boundary}
\end{equation}

\paragraph{Causal readout.}
Define the decay-scaled queries by
\begin{equation}
    (\mat{Q}_{\pi,[n]})_{j,:}
    =(\vct{q}_{\pi,[n]}^j)^\top
    =d_k^{-1/2}(\vct{\pi}_{[n]}^j\odot\vct{q}_{[n]}^j)^\top.
    \label{eq:chunk-query-scaling}
\end{equation}
For the prefix ending at position $j$, Equation~\eqref{eq:chunk-unrolled} gives
\begin{align}
    (\vct{o}_{[n]}^j)^\top
    &=d_k^{-1/2}(\vct{q}_{[n]}^j)^\top\mat{S}_{[n]}^j \notag\\
    &=(\vct{q}_{\pi,[n]}^j)^\top\widehat{\mat{S}}_{[n]}^j \notag\\
    &=(\vct{q}_{\pi,[n]}^j)^\top\mat{P}_{L,[n]}^{(j)}\mat{S}_{[n]}\mat{P}_{R,[n]}^{(j)} \notag\\
    &\quad+\sum_{i=1}^{j}\eta_{[n]}^i
    ((\vct{q}_{\pi,[n]}^j)^\top\vct{h}_{[n],i}^{(j)})(\vct{z}_{[n],i}^{(j)})^\top,
    \label{eq:causal-prefix-read}
\end{align}
where the transition products and propagated directions follow
Equations~\eqref{eq:chunk-unrolled}--\eqref{eq:chunk-effective-directions}
with endpoint $C$ replaced by $j$.
Denote the row-stacked first and second terms of the last expression by
$\mat{O}_{\mathrm{init},[n]}$ and $\mat{O}_{\mathrm{write},[n]}$, respectively.
Substituting the left-side WY identities from
Equations~\eqref{eq:chunk-transition-products} and~\eqref{eq:chunk-propagated-factors}
and stacking the causal reads gives
\begin{equation}
    \begin{aligned}
    \begin{bmatrix}
        (\vct{q}_{\pi,[n]}^1)^\top\mat{P}_{L,[n]}^{(1)}\\
        \vdots\\
        (\vct{q}_{\pi,[n]}^C)^\top\mat{P}_{L,[n]}^{(C)}
    \end{bmatrix}
    =\mat{Q}_{\pi,[n]}
    -\operatorname{tril}(\mat{Q}_{\pi,[n]}\bar{\mat{K}}_{[n]}^\top)
    \mat{M}_{L,[n]}\bar{\mat{U}}_{[n]}.
    \end{aligned}
    \label{eq:causal-left-prefix-expansion}
\end{equation}
Define
\begin{equation}
    \begin{aligned}
    \mat{G}_{L,[n]}
    &=\operatorname{tril}(\mat{Q}_{\pi,[n]}\bar{\mat{K}}_{[n]}^\top)
    \mat{M}_{L,[n]},\\
    \mat{Q}_{\mathrm{eff},[n]}
    &=\mat{Q}_{\pi,[n]}-\mat{G}_{L,[n]}\bar{\mat{U}}_{[n]},\\
    \mat{O}_{\mathrm{base},[n]}
    &=\mat{Q}_{\mathrm{eff},[n]}\mat{S}_{[n]}.
    \end{aligned}
    \label{eq:chunk-readout-factors}
\end{equation}
Substituting the right-side WY form into the initial-state and new-write
terms and retaining causal coefficients yields
\begin{align}
    \mat{O}_{\mathrm{init},[n]}
    &=\mat{O}_{\mathrm{base},[n]}
    -\operatorname{tril}(\mat{O}_{\mathrm{base},[n]}\mat{F}_{R,[n]}^\top)
    \hat{\mat{V}}_{[n]},
    \label{eq:causal-initial-contribution}\\
    \mat{O}_{\mathrm{write},[n]}
    &=\operatorname{tril}(\mat{G}_{L,[n]}\mat{N}_{[n]}\mat{M}_{R,[n]}^\top)
    \hat{\mat{V}}_{[n]}.
    \label{eq:causal-write-contribution}
\end{align}
Adding these contributions and defining
\begin{align}
    \mat{\Phi}_{[n]}&=\mat{G}_{L,[n]}\mat{N}_{[n]}\mat{M}_{R,[n]}^\top, \notag\\
    \mat{G}_{R,[n]}&=\operatorname{tril}\!\left(
    \mat{\Phi}_{[n]}-\mat{O}_{\mathrm{base},[n]}\mat{F}_{R,[n]}^\top
    \right)
    \label{eq:chunk-right-coefficients}
\end{align}
gives
\begin{align}
    \mat{O}_{[n]}
    &=\mat{O}_{\mathrm{init},[n]}+\mat{O}_{\mathrm{write},[n]} \notag\\
    &=\mat{O}_{\mathrm{base},[n]}+\operatorname{tril}\!\left(
    \mat{\Phi}_{[n]}-\mat{O}_{\mathrm{base},[n]}\mat{F}_{R,[n]}^\top\right)\hat{\mat{V}}_{[n]} \notag\\
    &=\mat{O}_{\mathrm{base},[n]}+\mat{G}_{R,[n]}\hat{\mat{V}}_{[n]}.
    \label{eq:chunk-output}
\end{align}
Here row $j$ of $\mat{O}_{[n]}$ is $(\vct{o}_{[n]}^{j})^\top$.
The operator $\operatorname{tril}$ retains the lower-triangular part, including
the diagonal, and sets all entries above the diagonal to zero.
Algorithm~\ref{alg:ddn-chunk} summarizes the computation, and
Appendix~\ref{app:implementation-details} gives the implementation details.

\begin{algorithm}[t]
\caption{DDN chunkwise computation}
\label{alg:ddn-chunk}
\footnotesize
\setlength{\abovedisplayskip}{2pt}
\setlength{\belowdisplayskip}{2pt}
\setlength{\abovedisplayshortskip}{2pt}
\setlength{\belowdisplayshortskip}{2pt}
\begin{algorithmic}[1]
\REQUIRE Incoming state $\mat{S}_{[n]}$; per-token $\vct{q}_{[n]}^{i}$,
$\vct{k}_{[n]}^{i}$, $\beta_{[n]}^{i}$, $\gamma_{[n]}^{i}$,
$\vct{v}_{[n]}^{i}$, $\hat{\vct{v}}_{[n]}^{i}$, $\vct{e}_{[n]}^{i}$,
$\tau_{[n]}^{i}$, and $\vct{\alpha}_{[n]}^{i}$ for $i=1,\ldots,C$
\ENSURE $\mat{O}_{[n]}$ and $\mat{S}_{[n+1]}$
\STATE \textbf{Normalize decay.} Compute $\vct{\pi}_{[n]}^{i}$ and
$\mat{\Pi}_{[n]}^{C}$ using Equation~\eqref{eq:chunk-cumulative-decay}.
Set $\eta_{[n]}^{i}=\beta_{[n]}^{i}\lVert \vct{v}_{[n]}^{i}\rVert_2
+\gamma_{[n]}^{i}(1-\beta_{[n]}^{i})\tau_{[n]}^{i}$.
\STATE \textbf{Triangular solution.} Stack
$((\mat{\Pi}_{[n]}^{i})^{-1}\vct{k}_{[n]}^{i})^\top$,
$\beta_{[n]}^{i}(\mat{\Pi}_{[n]}^{i}\vct{k}_{[n]}^{i})^\top$,
$(\vct{e}_{[n]}^{i})^\top$, and $(\hat{\vct{v}}_{[n]}^{i})^\top$ as rows of
$\bar{\mat{K}}_{[n]}$, $\bar{\mat{U}}_{[n]}$, $\mat{E}_{[n]}$, and $\hat{\mat{V}}_{[n]}$,
and set
$\mat{N}_{[n]}=\operatorname{Diag}(\eta_{[n]}^{1},\ldots,
\eta_{[n]}^{C})$ and
$\mat{\Gamma}_{[n]}=\operatorname{Diag}(\gamma_{[n]}^{1},\ldots,
\gamma_{[n]}^{C})$.
Compute $\mat{M}_{L,[n]}$, $\mat{M}_{R,[n]}$, $\mat{H}_{[n]}$, and $\mat{Z}_{[n]}$ using
Equations~\eqref{eq:chunk-erasure-coefficients}--\eqref{eq:chunk-propagated-factors}.
Form $\mat{Q}_{\pi,[n]}$ using Equation~\eqref{eq:chunk-query-scaling} and
compute $\mat{G}_{L,[n]}$ using Equation~\eqref{eq:chunk-readout-factors}.
\STATE \textbf{State-independent factors.}
\begin{equation*}
    \begin{gathered}
    \mat{F}_{R,[n]}\leftarrow \mat{M}_{R,[n]}\mat{\Gamma}_{[n]}\mat{E}_{[n]},\\
    \mat{Q}_{\mathrm{eff},[n]}\leftarrow
    \mat{Q}_{\pi,[n]}-\mat{G}_{L,[n]}\bar{\mat{U}}_{[n]},\\
    \mat{\Phi}_{[n]}\leftarrow \mat{G}_{L,[n]}\mat{N}_{[n]}\mat{M}_{R,[n]}^\top.
    \end{gathered}
\end{equation*}
\STATE \textbf{State update.}
\begin{equation*}
    \begin{gathered}
    \mat{R}_{[n]}\leftarrow
    \mat{S}_{[n]}-\mat{S}_{[n]}\mat{F}_{R,[n]}^\top\hat{\mat{V}}_{[n]},\\
    \mat{\Psi}_{[n]}\leftarrow \mat{N}_{[n]}\mat{Z}_{[n]}-\bar{\mat{U}}_{[n]}\mat{R}_{[n]},\\
    \mat{S}_{[n+1]}\leftarrow\mat{\Pi}_{[n]}^{C}
    (\mat{R}_{[n]}+\mat{H}_{[n]}\mat{\Psi}_{[n]}).
    \end{gathered}
\end{equation*}
\STATE \textbf{Causal readout.}
\begin{equation*}
    \begin{gathered}
    \mat{O}_{\mathrm{base},[n]}\leftarrow \mat{Q}_{\mathrm{eff},[n]}\mat{S}_{[n]},\\
    \mat{G}_{R,[n]}\leftarrow\operatorname{tril}\!\left(
    \mat{\Phi}_{[n]}-\mat{O}_{\mathrm{base},[n]}\mat{F}_{R,[n]}^\top\right),\\
    \mat{O}_{[n]}\leftarrow \mat{O}_{\mathrm{base},[n]}+\mat{G}_{R,[n]}\hat{\mat{V}}_{[n]}.
    \end{gathered}
\end{equation*}
\end{algorithmic}
\end{algorithm}

\section{Experiments}
\label{sec:experiments}

\paragraph{Setup.}
We compare DDN against KDA, GDN-2, and Mamba-3 to evaluate its effectiveness
among recent recurrent models. For Mamba-3, we evaluate both SISO and rank-4
MIMO variants. The 130M,
340M, and 1B models are trained on approximately 5B, 15B, and 50B FineWeb-Edu
tokens, respectively. Within each scale, we match total parameter counts by
adjusting model and MLP widths to compensate for differences in recurrent
parameters. Following the Mamba-3 and GDN-2
setups~\citep{lahoti2026mamba3,hatamizadeh2026gated}, we preserve the 
multi-value attention (MVA) head structure of Mamba-3, 
resulting in more heads than the linear-attention baselines.
This shape adjustment keeps the effective state size per layer
$n_{\mathrm{head}}d_kd_v$ identical across all baselines at each model scale. All runs use
AdamW, gradient clipping at 1.0, cosine
learning-rate decay, and a context length of 2,048. Further architecture,
optimization, and evaluation details are provided in
Appendix~\ref{app:experimental-details}.

\paragraph{Language modeling and zero-shot downstream tasks.}
Table~\ref{tab:main-results} reports FineWeb-Edu training perplexity,
WikiText-2 and LAMBADA perplexity, and zero-shot accuracy on the seven
downstream tasks from LAMBADA through OpenBookQA. In language modeling, DDN
achieves lower perplexity than KDA on all three measures at every scale and
ranks among the leading models in most comparisons. These results suggest that the
value-side delta rule improves language modeling. On zero-shot downstream
tasks, DDN achieves the highest average
downstream accuracy among all evaluated baselines at each scale.

\begin{table*}[p]
\centering
\caption{Results on language modeling and zero-shot downstream tasks at the 130M,
340M, and 1B scales after approximately 5B, 15B, and 50B FineWeb-Edu training tokens, respectively. All accuracy values
are percentages. Bold and underlined values denote
the best and second-best distinct scores within each model-size group in each column, respectively.}
\label{tab:main-results}
\resizebox{\linewidth}{!}{
\begin{tabular}{lccccccccccc}
    \toprule
    Model & FW-Edu & WT2 & LAMB. & LAMB. & PIQA & HellaS. &
    WinoGr. & ARC-E & ARC-C & OBQA & Avg. \\
    & train ppl $\downarrow$ & ppl $\downarrow$ & ppl $\downarrow$ & acc $\uparrow$ &
    acc $\uparrow$ & acc$_n$ $\uparrow$ & acc $\uparrow$ & acc $\uparrow$ &
    acc $\uparrow$ & acc $\uparrow$ & acc $\uparrow$ \\
    \midrule
    \multicolumn{12}{c}{130M models; 5B tokens} \\
    Mamba-3 SISO & 15.10 & 44.58 & 68.20 & 26.37 & \textbf{63.11} & 31.97 & 50.91 & 50.46 & \underline{20.39} & 18.80 & 37.43 \\
    Mamba-3 MIMO & 14.93 & 43.89 & 73.03 & 25.97 & 62.08 & 32.22 & \textbf{52.64} & \underline{51.01} & \textbf{20.73} & \textbf{19.60} & \underline{37.75} \\
    GDN-2 & 15.22 & 42.79 & \underline{67.88} & \underline{26.41} & 62.51 & \underline{32.28} & 51.70 & 50.93 & 20.14 & \underline{19.20} & 37.60 \\
    KDA & \underline{14.79} & \underline{42.53} & 69.91 & 25.69 & \underline{62.95} & 32.14 & 52.09 & 49.66 & 20.14 & 19.00 & 37.38 \\
    DDN (ours) & \textbf{14.78} & \textbf{42.36} & \textbf{60.93} & \textbf{27.75} & 62.35 & \textbf{32.49} & \underline{52.41} & \textbf{51.22} & \underline{20.39} & \textbf{19.60} & \textbf{38.03} \\
    \hline
    \multicolumn{12}{c}{340M models; 15B tokens} \\
    Mamba-3 SISO & 10.97 & 28.68 & 29.62 & 34.93 & 66.00 & \underline{40.46} & 49.96 & \textbf{59.18} & \underline{25.60} & 22.60 & 42.67 \\
    Mamba-3 MIMO & \textbf{10.81} & \underline{27.63} & \textbf{28.33} & 34.83 & \textbf{66.76} & \textbf{41.12} & 50.67 & \underline{59.09} & 24.92 & \underline{23.60} & \underline{43.00} \\
    GDN-2 & 11.19 & 29.40 & 29.18 & \textbf{36.35} & 65.51 & 39.48 & \textbf{52.25} & 56.06 & 23.21 & 20.20 & 41.87 \\
    KDA & 10.93 & 28.13 & 29.09 & \underline{35.92} & 65.83 & 40.31 & 51.38 & 57.53 & 24.23 & 21.60 & 42.40 \\
    DDN (ours) & \underline{10.88} & \textbf{27.43} & \underline{28.41} & \underline{35.92} & \underline{66.32} & 40.34 & \underline{51.46} & 58.33 & \textbf{27.30} & \textbf{24.40} & \textbf{43.44} \\
    \hline
    \multicolumn{12}{c}{1B models; 50B tokens} \\
    Mamba-3 SISO & 8.29 & 19.38 & 14.81 & 43.22 & 71.27 & \underline{53.00} & 54.93 & 67.00 & 32.42 & \underline{29.40} & 50.18 \\
    Mamba-3 MIMO & \textbf{8.20} & 19.10 & 15.14 & 42.75 & \underline{71.60} & \textbf{53.96} & \underline{56.20} & \textbf{68.01} & 32.42 & 28.40 & \underline{50.48} \\
    GDN-2 & 8.63 & 19.30 & 14.44 & \textbf{45.62} & 71.00 & 51.84 & \textbf{56.59} & 67.17 & 30.97 & 28.60 & 50.26 \\
    KDA & 8.23 & \underline{19.07} & \underline{13.86} & 44.29 & 71.22 & 52.84 & 56.12 & 66.67 & \textbf{33.28} & 28.20 & 50.37 \\
    DDN (ours) & \underline{8.22} & \textbf{19.00} & \textbf{13.56} & \underline{44.98} & \textbf{71.65} & 52.77 & 54.46 & \underline{67.59} & \underline{32.85} & \textbf{29.80} & \textbf{50.59} \\
    \bottomrule
\end{tabular}
}
\vspace{0.8em}
\centering
\caption{RULER-style needle retrieval accuracy in percent at the 130M,
340M, and 1B scales after approximately 5B, 15B, and 50B training tokens, with 500 examples per task. Bold and underlined values denote the best and second-best distinct scores
within each model-size group in each column.}
\label{tab:niah-results}
\resizebox{\linewidth}{!}{
\begin{tabular}{lccc ccc ccc ccc c}
    \toprule
    & \multicolumn{3}{c}{NIAH-Single-1}
    & \multicolumn{3}{c}{NIAH-Single-2}
    & \multicolumn{3}{c}{NIAH-Single-3}
    & \multicolumn{3}{c}{NIAH-MultiKey-1} & \\
    Model & 1024 & 2048 & 4096 & 1024 & 2048 & 4096 &
    1024 & 2048 & 4096 & 1024 & 2048 & 4096 & Avg. \\
    \midrule
    \multicolumn{14}{c}{130M models; 5B tokens} \\
    Mamba-3 SISO & 22.2 & 6.4 & 3.4 & 66.2 & 18.6 & 10.0 & 0.0 & 0.0 & \underline{0.0} & 9.0 & 10.4 & 9.2 & 12.95 \\
    Mamba-3 MIMO & 41.6 & 14.4 & 5.2 & 41.0 & 9.6 & 3.6 & 0.0 & 0.0 & \underline{0.0} & 9.6 & 10.0 & 4.8 & 11.65 \\
    GDN-2 & 68.8 & 29.6 & 14.2 & 82.8 & \underline{34.0} & \textbf{28.4} & \underline{1.8} & \underline{0.2} & \textbf{0.2} & \underline{23.8} & \textbf{22.0} & \underline{16.8} & 26.88 \\
    KDA & \underline{99.0} & \underline{96.8} & \underline{52.8} & \textbf{93.0} & 33.6 & \underline{21.0} & 0.4 & \underline{0.2} & \underline{0.0} & \textbf{25.4} & \underline{21.4} & 13.6 & \underline{38.10} \\
    DDN (ours) & \textbf{99.2} & \textbf{97.4} & \textbf{82.8} & \underline{84.6} & \textbf{36.0} & 17.2 & \textbf{5.4} & \textbf{2.4} & \textbf{0.2} & 22.6 & 21.0 & \textbf{18.0} & \textbf{40.57} \\
    \hline
    \multicolumn{14}{c}{340M models; 15B tokens} \\
    Mamba-3 SISO & 86.2 & 36.6 & 19.8 & 98.2 & 86.0 & 32.0 & 7.0 & 3.6 & 2.0 & 16.0 & 16.6 & 16.2 & 35.02 \\
    Mamba-3 MIMO & 94.0 & 66.2 & 28.0 & 96.4 & 90.0 & 35.6 & 17.0 & 8.4 & \underline{2.2} & 21.6 & 21.4 & 17.6 & 41.53 \\
    GDN-2 & \underline{99.8} & 96.6 & \underline{65.2} & \textbf{100.0} & \textbf{98.4} & \textbf{48.8} & 2.0 & 4.0 & \underline{2.2} & 22.6 & 22.4 & \textbf{20.8} & 48.57 \\
    KDA & \textbf{100.0} & \underline{98.8} & 64.2 & \underline{99.2} & \underline{95.6} & \underline{40.6} & \underline{27.8} & \underline{13.6} & 2.0 & \textbf{27.4} & \textbf{25.4} & \underline{19.8} & \underline{51.20} \\
    DDN (ours) & \textbf{100.0} & \textbf{100.0} & \textbf{100.0} & 98.4 & 67.2 & 14.4 & \textbf{58.0} & \textbf{37.2} & \textbf{7.0} & \underline{23.6} & \underline{23.2} & 17.4 & \textbf{53.87} \\
    \hline
    \multicolumn{14}{c}{1B models; 50B tokens} \\
    Mamba-3 SISO & 98.6 & 75.4 & 16.6 & 97.2 & 50.2 & 24.8 & 68.4 & 28.8 & 11.0 & \underline{48.8} & \underline{35.0} & 12.2 & 47.25 \\
    Mamba-3 MIMO & \underline{99.8} & 83.0 & 43.2 & \textbf{100.0} & \underline{97.4} & 56.4 & 83.4 & 50.8 & 7.0 & \textbf{53.2} & \textbf{35.4} & 19.2 & 60.73 \\
    GDN-2 & \textbf{100.0} & \textbf{100.0} & \underline{99.8} & \underline{99.8} & 95.4 & \underline{56.6} & \underline{94.2} & \underline{78.2} & \underline{29.4} & 30.6 & 25.8 & \underline{23.0} & \underline{69.40} \\
    KDA & \textbf{100.0} & \underline{97.6} & 96.4 & 92.8 & \textbf{100.0} & 54.2 & 59.4 & 67.2 & 19.0 & 34.4 & 30.2 & 20.0 & 64.27 \\
    DDN (ours) & \textbf{100.0} & \textbf{100.0} & \textbf{100.0} & \textbf{100.0} & \textbf{100.0} & \textbf{80.8} & \textbf{97.0} & \textbf{83.6} & \textbf{36.0} & 37.6 & 32.2 & \textbf{27.0} & \textbf{74.52} \\
    \bottomrule
\end{tabular}
}
\end{table*}

\begin{table}[htbp]
\centering
\caption{Real-world retrieval at the 130M, 340M, and 1B scales after
approximately 5B, 15B, and 50B training tokens. 
Bold and underlined values denote the best and second-best distinct scores
within each model-size group in each column.}
\label{tab:retrieval-results}
\setlength{\tabcolsep}{2.5pt}
\renewcommand{\arraystretch}{1.05}
\resizebox{\linewidth}{!}{%
\begin{tabular}{lrrrrrrr}
    \toprule
    Model & SWDE & SQuAD & FDA & TQA & NQ & DROP & Avg. \\
    \midrule
    \multicolumn{8}{c}{130M models; 5B tokens} \\
    Mamba-3 SISO & 5.85 & 18.97 & 0.82 & 35.13 & 4.47 & 11.60 & 12.81 \\
    Mamba-3 MIMO & 6.57 & 16.89 & 1.36 & 34.83 & 4.72 & 13.51 & 12.98 \\
    GDN-2 & \underline{10.35} & \underline{21.35} & \underline{2.09} & 37.03 & 5.61 & \textbf{16.77} & \underline{15.53} \\
    KDA & 8.82 & 20.07 & 1.91 & \underline{37.92} & \underline{5.70} & \underline{16.68} & 15.18 \\
    DDN (ours) & \textbf{10.62} & \textbf{21.95} & \textbf{2.63} & \textbf{39.40} & \textbf{5.96} & 16.53 & \textbf{16.18} \\
    \hline
    \multicolumn{8}{c}{340M models; 15B tokens} \\
    Mamba-3 SISO & 14.40 & 27.71 & 3.27 & 50.24 & 8.46 & 19.26 & 20.56 \\
    Mamba-3 MIMO & 15.84 & 29.89 & 2.72 & \textbf{51.72} & 9.00 & \underline{19.50} & 21.45 \\
    GDN-2 & \underline{17.10} & \textbf{30.80} & 3.63 & 50.42 & 9.34 & 17.97 & \underline{21.54} \\
    KDA & 16.83 & \underline{30.23} & \underline{3.81} & 49.82 & \textbf{9.60} & 17.54 & 21.31 \\
    DDN (ours) & \textbf{18.45} & 29.89 & \textbf{8.71} & \underline{51.13} & \underline{9.57} & \textbf{21.56} & \textbf{23.22} \\
    \hline
    \multicolumn{8}{c}{1B models; 50B tokens} \\
    Mamba-3 SISO & 27.18 & 35.29 & 6.08 & 58.83 & 15.24 & 23.19 & 27.63 \\
    Mamba-3 MIMO & 31.86 & 36.03 & 11.16 & 60.19 & 15.33 & 22.14 & 29.45 \\
    GDN-2 & 31.95 & \textbf{39.21} & \textbf{17.42} & \textbf{62.97} & \underline{17.01} & 22.23 & \textbf{31.80} \\
    KDA & \textbf{35.46} & \underline{37.80} & 10.62 & 61.08 & 15.30 & \textbf{25.11} & 30.90 \\
    DDN (ours) & \underline{32.58} & 37.06 & \underline{12.89} & \underline{62.50} & \textbf{17.45} & \underline{24.72} & \underline{31.20} \\
    \bottomrule
\end{tabular}
}
\end{table}

\paragraph{Synthetic NIAH tasks.}
Table~\ref{tab:niah-results} evaluates four needle variants from
RULER~\citep{hsieh2024ruler} at context lengths from 1,024 to 4,096.
DDN achieves the highest average score at all three model scales. At 1B, it
achieves the highest accuracy across all NIAH-Single settings, with particularly
pronounced gains on NIAH-Single-2 at 4,096 tokens and NIAH-Single-3. 
These results suggest that the state management provided by the
value-side delta rule substantially improves associative recall.

\paragraph{Real-world retrieval tasks.}
Table~\ref{tab:retrieval-results} reports six
recall-intensive real-world tasks from Just Read Twice
\citep{arora2024justreadtwice}, with inputs truncated to a maximum of 2,048
tokens. DDN outperforms KDA in average accuracy at all
three scales, achieving the highest average at 130M and 340M. At 1B, it ranks
second with 31.20\%, slightly below GDN-2 at 31.80\%. GDN-2 shows a
particularly large advantage on FDA at this scale, scoring 17.42\% compared
with 12.89\% for DDN and 10.62\% for KDA. These task-dependent gains suggest
that GDN-2 and DDN may improve different aspects of recall.

\begin{figure}[htbp]
\centering
\includegraphics[width=0.8\columnwidth]{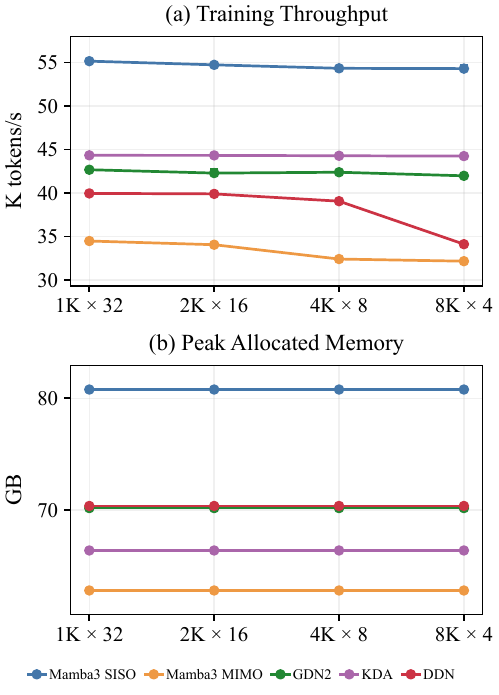}
\caption{Training throughput and peak allocated memory of 1B models on an
NVIDIA H200. The horizontal axis denotes sequence length $\times$
batch size. All baselines retain their default recomputation settings.}
\label{fig:training-efficiency}
\end{figure}

\paragraph{Training and inference efficiency.}
Figure~\ref{fig:training-efficiency} compares end-to-end BF16 training
throughput and peak allocated memory of 1B models on an NVIDIA H200.
All baselines retain the default recomputation settings of their original
implementations.
DDN's throughput is modestly lower than KDA's and GDN-2's, and higher than
Mamba-3 MIMO's. Mamba-3 SISO saves recurrent states and other forward
intermediates for backward, favoring higher throughput at the cost of greater
memory usage. DDN's peak allocated memory is slightly higher than KDA's
and comparable to GDN-2's. Both DDN and Mamba-3 MIMO show reduced
throughput at $8\mathrm{K}\times4$. A possible reason is that the current
implementations provide limited parallelism at small batch sizes, as discussed
in Appendix~\ref{app:training-parallelism}.
For inference, Table~\ref{tab:inference-latency} reports prefill
and prefill-plus-decode latency for the 1B models on an NVIDIA H200 with
batch size 64. 
The generation length equals the input length. 
GDN-2, DDN, and Mamba-3 MIMO have higher prefill and
prefill-plus-decode latencies than Mamba-3 SISO and KDA. DDN's latencies are
close to those of GDN-2 and Mamba-3 MIMO.

\begin{table}[!ht]
\centering
\footnotesize
\setlength{\tabcolsep}{3pt}
\renewcommand{\arraystretch}{1.1}
\caption{Prefill and prefill + decode latency in seconds for 1B models on an
NVIDIA H200 with batch size 64. Generation length equals input length.
Bold and underlined values denote the lowest and second-lowest
distinct latencies in each column within each group, respectively.}
\label{tab:inference-latency}
\begin{tabular}{@{}lrrrr@{}}
\toprule
Model & 1024 & 2048 & 4096 & 8192 \\
\midrule
\multicolumn{5}{c}{Prefill} \\
Mamba-3 SISO & \textbf{0.34} & \textbf{0.68} & \textbf{1.35} & \textbf{2.69} \\
Mamba-3 MIMO & 0.40 & 0.81 & 1.64 & 3.35 \\
GDN-2 & 0.40 & 0.80 & 1.58 & 3.16 \\
KDA & \underline{0.35} & \underline{0.70} & \underline{1.39} & \underline{2.81} \\
DDN (ours) & 0.39 & 0.79 & 1.57 & 3.15 \\
\hline
\multicolumn{5}{c}{Prefill + Decode} \\
Mamba-3 SISO & \textbf{3.65} & \textbf{7.31} & \textbf{14.66} & \textbf{27.44} \\
Mamba-3 MIMO & 4.16 & 8.34 & 16.70 & 33.46 \\
GDN-2 & \underline{4.13} & \underline{8.26} & \underline{16.53} & 33.11 \\
KDA & \textbf{3.65} & \textbf{7.31} & \textbf{14.66} & \underline{29.34} \\
DDN (ours) & 4.18 & 8.35 & 16.68 & 33.37 \\
\bottomrule
\end{tabular}
\end{table}

\paragraph{Ablation study.}
To isolate the role of the learned channel gate, we train 130M and 340M DDN
variants with $\vct{e}_t=\hat{\vct{v}}_t$ for approximately 5B and 15B tokens,
respectively. Table~\ref{tab:ablation-summary} summarizes the
results. At both scales, the gated and ungated DDN variants improve retrieval
and NIAH accuracy over KDA, supporting the effectiveness of the value-side delta rule.
The gated variant achieves stronger overall performance, suggesting that
selecting relevant value features benefits value-side address reorganization.

\begin{table}[!ht]
\centering
\caption{Ablation results for 130M and 340M models trained on approximately 5B
and 15B FineWeb-Edu tokens, respectively. Downstream, Retrieval, and NIAH
report average accuracy in percent. Bold and underlined values
denote the best and second-best scores at each scale.}
\label{tab:ablation-summary}
\setlength{\tabcolsep}{3pt}
\renewcommand{\arraystretch}{1.05}
\begin{tabular}{lccc}
\toprule
Model & Downstream & Retrieval & NIAH \\
\midrule
\multicolumn{4}{c}{130M models; 5B tokens} \\
KDA & 37.38 & 15.18 & 38.10 \\
DDN ($\vct{e}_t=\hat{\vct{v}}_t$) & \underline{37.41} & \underline{15.87} & \underline{39.45} \\
DDN & \textbf{38.03} & \textbf{16.18} & \textbf{40.57} \\
\midrule
\multicolumn{4}{c}{340M models; 15B tokens} \\
KDA & \underline{42.40} & 21.30 & 51.20 \\
DDN ($\vct{e}_t=\hat{\vct{v}}_t$) & 42.16 & \textbf{23.48} & \underline{52.42} \\
DDN & \textbf{43.44} & \underline{23.22} & \textbf{53.87} \\
\bottomrule
\end{tabular}
\end{table}

\section{Conclusion}

We presented Dual DeltaNet (DDN), which extends KDA with complementary
key-side and value-side delta rules. 
The key-side update corrects the value associated with the current
key, while the value-side update utilizes selected value features to reorganize
related content across addresses. A chunkwise parallel formulation and a
rank-2 recurrent form support efficient training and inference with modest
overhead relative to KDA. Experiments across the 130M, 340M, and 1B scales
show improvements over KDA in language modeling, zero-shot downstream tasks,
and retrieval, with particularly pronounced gains on the RULER
needle-in-a-haystack benchmark. These results suggest that key-side
associative correction and value-side address reorganization can
jointly improve the management of fixed-size recurrent memory.

\bibliography{reference}
\bibliographystyle{icml2025}

\appendix
\section{Compact WY Representation}
\label{app:compact-wy}

\begin{lemma}[Compact WY form of rank-one transitions]
\label{lem:compact-wy}
Let $\vct{u}_i,\vct{w}_i\in\mathbb R^d$ and $\mat{Q}_i=\mat{I}-\vct{u}_i\vct{w}_i^\top$ for
$i=1,\ldots,m$. Set $\mat{U}=[\vct{u}_1,\ldots,\vct{u}_m]$,
$\mat{W}=[\vct{w}_1,\ldots,\vct{w}_m]$, and
\begin{align}
    \mat{T}&=\operatorname{tril}(\mat{W}^\top \mat{U},-1),
    \qquad \mat{M}=(\mat{I}+\mat{T})^{-1}.
    \label{eq:wy-coefficients}
\end{align}
Thus $T_{ij}=\vct{w}_i^\top \vct{u}_j$ for $j<i$, and $T_{ij}=0$ otherwise.
The inverse exists for arbitrary $\vct{u}_i,\vct{w}_i$, since $\mat{I}+\mat{T}$ is unit
lower triangular. Define the propagated columns
$\vct{f}_i=\mat{Q}_m\cdots \mat{Q}_{i+1}\vct{u}_i$ and $\mat{F}=[\vct{f}_1,\ldots,\vct{f}_m]$,
with empty products equal to $\mat{I}$. Then
\begin{equation}
    \mat{F}(\mat{I}+\mat{T})=\mat{U},\qquad \mat{F}=\mat{U}\mat{M}.
    \label{eq:wy-propagated-factors}
\end{equation}
The transition product admits the compact representation
\begin{equation}
    \mat{Q}_m\cdots \mat{Q}_1=\mat{I}-\mat{F}\mat{W}^\top=\mat{I}-\mat{U}\mat{M}\mat{W}^\top.
    \label{eq:wy-product}
\end{equation}
\end{lemma}

\begin{proof}
The cancellation can be seen by expanding the successive differences:
\begin{align*}
    \mat{Q}_m\cdots \mat{Q}_{i+1}-\mat{I}
    &=(\mat{Q}_m\cdots \mat{Q}_{i+1}-\mat{Q}_m\cdots \mat{Q}_{i+2})\\
    &\quad+(\mat{Q}_m\cdots \mat{Q}_{i+2}-\mat{Q}_m\cdots \mat{Q}_{i+3})\\
    &\quad+\cdots+(\mat{Q}_m-\mat{I}).
\end{align*}
Thus, for $0\leq i\leq m$, with an empty sum when $i=m$,
\begin{align}
    \mat{Q}_m\cdots \mat{Q}_{i+1}
    &=\mat{I}+\sum_{j=i+1}^{m}\bigl(\mat{Q}_m\cdots \mat{Q}_j-\mat{Q}_m\cdots \mat{Q}_{j+1}\bigr) \notag\\
    &=\mat{I}+\sum_{j=i+1}^{m}(\mat{Q}_m\cdots \mat{Q}_{j+1})(\mat{Q}_j-\mat{I}) \notag\\
    &=\mat{I}-\sum_{j=i+1}^{m}(\mat{Q}_m\cdots \mat{Q}_{j+1}\vct{u}_j)\vct{w}_j^\top \notag\\
    &=\mat{I}-\sum_{j=i+1}^{m}\vct{f}_j\vct{w}_j^\top,
    \label{eq:wy-telescoping}
\end{align}
where the third line uses $\mat{Q}_j-\mat{I}=-\vct{u}_j\vct{w}_j^\top$, and the last uses the
definition of $\vct{f}_j$.
For $i\geq1$, multiplying by $\vct{u}_i$ yields
$\vct{f}_i=\vct{u}_i-\sum_{j>i}\vct{f}_j(\vct{w}_j^\top \vct{u}_i)$.
These are the column equations of $\mat{F}(\mat{I}+\mat{T})=\mat{U}$, giving
$\mat{F}=\mat{U}\mat{M}$. Taking $i=0$ in Equation~\eqref{eq:wy-telescoping}
gives the product identity.
\end{proof}

\section{Implementation Details}
\label{app:implementation-details}

All implementations are built on the Flash Linear Attention (FLA)
library~\citep{yang2024fla}.

\paragraph{Decay parameterization.}
The diagonal decay is implemented as $\mat{D}_t=\operatorname{Diag}(\vct{\alpha}_t)$ with
\begin{equation}
    \vct{\alpha}_t
    =
    \exp(\vct{g}_t),
    \qquad
    \vct{g}_t
    =
    -\exp(a_{\mathrm{decay}})\odot\vct{\Delta}_t,
\end{equation}
where
\begin{equation}
    \vct{\Delta}_t
    =
    \operatorname{softplus}(\mat{W}_\Delta \vct{x}_t+\vct{b}_\Delta).
\end{equation}
Here $a_{\mathrm{decay}}$ is a learned per-head parameter and $\vct{\Delta}_t$ is a
positive input-dependent step size. This parameterization keeps
$\vct{g}_t<0$ and therefore $\vct{\alpha}_t\in(0,1)$.

\paragraph{State transition implementation.}
\label{app:training-parallelism}
State transition kernels for KDA and GDN-2 exploit the independence of state
columns through value tiling, assigning a separate program to each sequence,
attention head, and value tile. This increases parallelism beyond
one program per sequence and head. DDN's value-side read couples these
columns through a reduction over $d_v$, so value tiles cannot be processed
independently under the same scheme. Distributing them across programs would
require cross-program reductions and synchronization. We therefore assign
one program to each sequence and attention head, with chunks processed
sequentially within each program. We implement this strategy in
TileLang~\citep{wang2025tilelang}, keeping the full state resident in shared
memory. This strategy avoids cross-program
reductions, but limits the number of independently scheduled programs to
the batch size times the number of heads. At small batch sizes, the resulting
parallelism may be insufficient to fully utilize the GPU, contributing to
the throughput decrease at $8\mathrm{K}\times4$ in
Figure~\ref{fig:training-efficiency}.
The released Mamba-3 MIMO implementation similarly processes chunks
sequentially within one program per sequence and head~\citep{lahoti2026mamba3}.

\paragraph{Precision and recomputation.}
Most matrix operands and the carried recurrent state are stored in BF16,
with FP32 accumulation for matrix products. Normalization, cumulative
log-decay calculations, and triangular-solve intermediates are computed in
FP32. During training, we retain $\vct{q}$, $\vct{k}$, and
$\hat{\vct{v}}$, together with the inverse L2 norms computed during
normalization. The triangular inverses $\mat{M}_{L,[n]}$ and
$\mat{M}_{R,[n]}$ and the causal query--key interaction matrix
$\operatorname{tril}(\mat{Q}_{\pi,[n]}\bar{\mat{K}}_{[n]}^\top)$ are also
cached. Other state-transition and readout intermediates are recomputed
during backward to reduce activation memory.

\section{Experimental Details}
\label{app:experimental-details}

\paragraph{Architecture and parameter matching.}
Table~\ref{tab:model-configurations} summarizes the model configurations.
KDA, GDN-2, and DDN use 128-dimensional keys and values. Mamba-3 retains
its multi-value structure with expansion factor 2, one shared state group, and
state and head dimensions of 64. Consequently, every model at a
given scale has the same effective state size $n_{\mathrm{head}}d_kd_v$.
Model and MLP widths are adjusted within each scale
to compensate for differences in recurrent parameters.
In DDN, the decay and selector-gate logits are generated by separate low-rank
projections with bottleneck width $d_v$, following KDA's decay-gate parameterization.
We initialize all entries of $\vct{m}_t$ to $0.5$ and $\gamma_t$ to $0.05$,
giving $\vct{e}_t=\hat{\vct{v}}_t$ at initialization.

\begin{table}[!ht]
\centering
\caption{Architecture configurations. For the linear-attention models, $d_k$ and $d_v$ are the key and value
dimensions. For Mamba-3, they denote the state and head dimensions,
respectively.}
\label{tab:model-configurations}
\setlength{\tabcolsep}{5pt}
\renewcommand{\arraystretch}{1.05}
\resizebox{\columnwidth}{!}{%
\begin{tabular}{lrrrrrrr}
\toprule
Model & Depth & Heads & $d_k$ & $d_v$ & $d_{\mathrm{model}}$ &
$d_{\mathrm{MLP}}$ & State size \\
\midrule
\multicolumn{8}{c}{130M scale} \\
Mamba-3 SISO & 12 & 24 & 64 & 64 & 768 & 2,208 & 98,304 \\
Mamba-3 MIMO & 12 & 24 & 64 & 64 & 768 & 2,064 & 98,304 \\
GDN-2 & 12 & 6 & 128 & 128 & 768 & 2,112 & 98,304 \\
KDA & 12 & 6 & 128 & 128 & 816 & 2,336 & 98,304 \\
DDN (ours) & 12 & 6 & 128 & 128 & 800 & 2,320 & 98,304 \\
\midrule
\multicolumn{8}{c}{340M scale} \\
Mamba-3 SISO & 24 & 32 & 64 & 64 & 1,024 & 2,048 & 131,072 \\
Mamba-3 MIMO & 24 & 32 & 64 & 64 & 1,024 & 1,920 & 131,072 \\
GDN-2 & 24 & 8 & 128 & 128 & 904 & 2,560 & 131,072 \\
KDA & 24 & 8 & 128 & 128 & 992 & 2,816 & 131,072 \\
DDN (ours) & 24 & 8 & 128 & 128 & 976 & 2,800 & 131,072 \\
\midrule
\multicolumn{8}{c}{1B scale} \\
Mamba-3 SISO & 24 & 56 & 64 & 64 & 1,792 & 3,584 & 229,376 \\
Mamba-3 MIMO & 24 & 56 & 64 & 64 & 1,792 & 3,456 & 229,376 \\
GDN-2 & 24 & 14 & 128 & 128 & 1,616 & 4,352 & 229,376 \\
KDA & 24 & 14 & 128 & 128 & 1,760 & 4,864 & 229,376 \\
DDN (ours) & 24 & 14 & 128 & 128 & 1,744 & 4,848 & 229,376 \\
\bottomrule
\end{tabular}
}
\end{table}

\paragraph{Optimization.}
All models are trained on FineWeb-Edu with a batch of 256 sequences and
a sequence length of 2,048. All models share the same LLaMA tokenizer with
a vocabulary of 32,000 tokens for training and evaluation.
We utilize AdamW with $\beta_1=0.9$, $\beta_2=0.95$,
$\epsilon=10^{-8}$, weight decay 0.01, and gradient clipping at 1.0. The
learning rate follows a cosine schedule that decays to 10\% of its peak value.
Table~\ref{tab:training-details} summarizes the training steps,
peak learning rates, and warmup steps for all model scales.

\begin{table}[!ht]
\centering
\caption{Training steps, peak learning rates, and warmup steps of pretraining.}
\label{tab:training-details}
\setlength{\tabcolsep}{5pt}
\begin{tabular}{lccc}
    \toprule
    Scale & 130M & 340M & 1B \\
    \midrule
    Training steps & 9,500 & 30,000 & 95,000 \\
    Peak learning rate & $6\times10^{-4}$ & $4\times10^{-4}$ & $4\times10^{-4}$ \\
    Warmup steps & 512 & 1,024 & 1,024 \\
    \bottomrule
\end{tabular}
\end{table}

\paragraph{Evaluation protocol.}
All downstream and retrieval evaluations are zero-shot and are conducted
with \texttt{lm-evaluation-harness}~\citep{gao2021framework}. 
Responses are generated via greedy decoding, with a maximum of
48 new tokens for real-world retrieval and 128 for NIAH.

\paragraph{Language modeling.}
FineWeb-Edu training perplexity is the exponential of the mean logged loss
over the final 500 steps, sampled every 10 steps. We report word-level
perplexity on WikiText-2~\citep{merity2016pointer} and both perplexity and
final-word prediction accuracy on the OpenAI version of
LAMBADA~\citep{paperno2016lambada}.

\paragraph{Zero-shot downstream tasks.}
The seven tasks cover contextual word prediction, commonsense reasoning,
coreference resolution, and science question answering. 
We report raw accuracy for LAMBADA~\citep{paperno2016lambada},
PIQA~\citep{bisk2020piqa}, WinoGrande~\citep{sakaguchi2021winogrande},
ARC-Easy and ARC-Challenge~\citep{clark2018think}, and
OpenBookQA~\citep{mihaylov2018can}, and length-normalized accuracy for
HellaSwag~\citep{zellers2019hellaswag}.

\paragraph{Real-world retrieval tasks.}
We evaluate six recall-intensive tasks following prior work
~\citep{arora2024simple,arora2024justreadtwice}:
SWDE~\citep{hao2011one}, SQuAD Completion~\citep{rajpurkar2016squad},
FDA~\citep{arora2024simple}, TriviaQA Cloze~\citep{joshi2017triviaqa},
Natural Questions Cloze~\citep{kwiatkowski2019natural}, and
DROP Cloze~\citep{dua2019drop}.
Inputs are limited to 2,048 tokens. We report contains accuracy, counting a
response as correct when it contains any accepted answer string.

\paragraph{Synthetic NIAH tasks.}
The four RULER variants~\citep{hsieh2024ruler} test word-to-number retrieval
in repeated text (Single-1) or essays (Single-2), word-to-UUID retrieval in
essays (Single-3), and retrieval among four word-to-number associations in
essays (MultiKey-1). Each variant has 500 examples at each context length
of 1,024, 2,048, and 4,096 tokens.

\end{document}